\documentclass[sigconf]{acmart}
\AtBeginDocument{%
  \providecommand\BibTeX{{%
    \normalfont B\kern-0.5em{\scshape i\kern-0.25em b}\kern-0.8em\TeX}}}

\usepackage{balance}
\usepackage{makecell}

\usepackage[capitalize]{cleveref}
\crefname{section}{Sec.}{Secs.}
\Crefname{section}{Section}{Sections}
\crefname{table}{Tab.}{Tabs.}
\Crefname{table}{Table}{Tables}
\crefname{figure}{Fig.}{Figs.}
\Crefname{figure}{Figure}{Figures}
\crefname{equation}{Eq.}{Eqs.}
\Crefname{equation}{Equation}{Equations}

\def\eg{\emph{e.g}.} 
\def\ie{\emph{i.e}.} 
\def\etc{\emph{etc}.}

\setcopyright{acmlicensed}
\copyrightyear{2023}
\acmYear{2023}
\acmDOI{10.1145/3581783.3611948}

\acmConference[MM '23] {Proceedings of the 31st ACM International Conference on Multimedia}{October 29--November 3, 2023}{Ottawa, ON, Canada.}
\acmBooktitle{Proceedings of the 31st ACM International Conference on Multimedia (MM '23), October 29--November 3, 2023, Ottawa, ON, Canada}
\acmPrice{15.00}
\acmISBN{979-8-4007-0108-5/23/10}

\acmSubmissionID{1130}

\begin{document}

\title{DUSA: Decoupled Unsupervised Sim2Real Adaptation for Vehicle-to-Everything Collaborative Perception}


\author{Xianghao Kong}
\affiliation{%
  \institution{Institute of Artificial Intelligence, Beihang University}
  \city{Beijing}
  \country{China}}
\email{refkxh@buaa.edu.cn}

\author{Wentao Jiang}
\affiliation{%
  \institution{Institute of Artificial Intelligence, Beihang University}
  \city{Beijing}
  \country{China}
}
\email{jiangwentao@buaa.edu.cn}

\author{Jinrang Jia}
\affiliation{%
 \institution{Baidu Inc.}
 \city{Beijing}
 \country{China}}
\email{jiajinrang@baidu.com}

\author{Yifeng Shi}
\authornote{Corresponding author.}
\affiliation{%
  \institution{Baidu Inc.}
  \city{Beijing}
  \country{China}}
\email{shiyifeng@baidu.com}

\author{Runsheng Xu}
\affiliation{%
  \institution{University of California, Los Angeles}
  \city{Los Angeles}
  \state{California}
  \country{USA}
  }
\email{rxx3386@ucla.edu}

\author{Si Liu}
\affiliation{%
  \institution{Institute of Artificial Intelligence, Beihang University}
  \city{Beijing}
  \country{China}}
\email{liusi@buaa.edu.cn}


\renewcommand{\shortauthors}{Xianghao Kong et al.}

\begin{abstract}
Vehicle-to-Everything (V2X) collaborative perception is crucial for the advancement of autonomous driving. 
However, achieving high-precision V2X perception requires a significant amount of annotated real-world data, which can always be expensive and hard to acquire.
Simulated data have raised much attention since they can be massively produced at an extremely low cost.
Nevertheless, the significant domain gap between simulated and real-world data, including differences in sensor type, reflectance patterns, and road surroundings, often leads to poor performance of models trained on simulated data when evaluated on real-world data.
In addition, there remains a domain gap between real-world collaborative agents, \eg ~different types of sensors may be installed on autonomous vehicles and roadside infrastructures with different extrinsics, further increasing the difficulty of sim2real generalization.
To take full advantage of simulated data, we present a new unsupervised sim2real domain adaptation method for V2X collaborative detection named Decoupled Unsupervised Sim2Real Adaptation (DUSA).
Our new method decouples the V2X collaborative sim2real domain adaptation problem into two sub-problems: sim2real adaptation and inter-agent adaptation.
For sim2real adaptation, we design a Location-adaptive Sim2Real Adapter (LSA) module to adaptively aggregate features from critical locations of the feature map and align the features between simulated data and real-world data via a sim/real discriminator on the aggregated global feature.
For inter-agent adaptation, we further devise a Confidence-aware Inter-agent Adapter (CIA) module to align the fine-grained features from heterogeneous agents under the guidance of agent-wise confidence maps.
Experiments demonstrate the effectiveness of the proposed DUSA approach on unsupervised sim2real adaptation from the simulated V2XSet dataset to the real-world DAIR-V2X-C dataset.
\end{abstract}





\maketitle

\section{Introduction}
Autonomous driving has become a rapidly developing subject in recent years. 
It has the potential of revolutionizing transportation systems by achieving better road safety and traffic efficiency. 
For autonomous driving, perception plays an essential part since it provides information about the surroundings for autonomous planning and control.
With the advancement of perception in autonomous driving, Vehicle-to-Everything (V2X) collaborative detection \cite{cooper, fcooper, who2com, when2com, v2vnet, disconet, fpvrcnn, opv2v, v2xvit, dair, crcnet} has received increasingly more attention in both academia and industry. 
The interaction between multiple agents could effectively compensate for the deficiencies of individual perception, \eg ~limited perception range and obstacle occlusion, prominently improving the perception ability of the ego vehicle.

Despite the benefits brought by V2X collaborative perception, training a high-precision V2X detection model applicable to autonomous driving systems requires a large amount of annotated real-world data, the acquisition of which can be both time-consuming and expensive. 
Collecting such data requires well-calibrated sensors which are placed on specialized vehicles and roadside infrastructure across various geographical locations. The process of synchronizing, designing the scenarios, and annotating raw data can be very labor-intensive.
In contrast, generating collaborative data for autonomous driving in simulation is much more cost-effective and efficient.
With the help of co-simulation tools such as CARLA \cite{carla}, SUMO \cite{sumo}, and OpenCDA \cite{opencda}, large-scale simulated collaborative perception datasets can be built without the need for human annotation.
Consequently, the construction of these datasets becomes significantly more manageable and accessible. 

\begin{figure}[t]
  \centering
  \includegraphics[width=\linewidth]{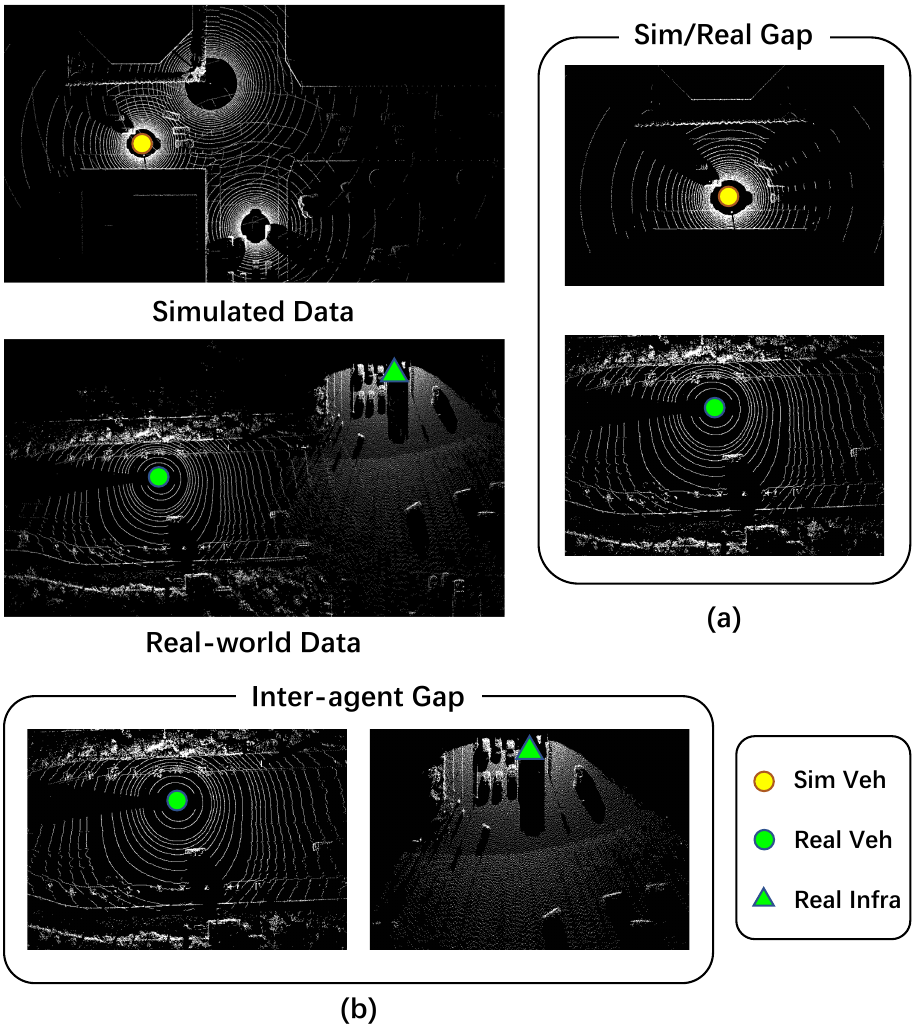}
  \caption{Domain gaps in V2X collaborative sim2real adaptation. We compare one sample from a simulated dataset V2XSet \cite{v2xvit} and another sample from a real-world dataset DAIR-V2X-C \cite{dair}. (a) Sim/real domain gap results in point clouds which differ in density, distribution, reflectance, \etc ~(b) Inter-agent domain gap caused by different LiDAR types and placement also results in distinct point clouds.}
  \Description{Sim/Real domain gap and Inter-agent domain gap.}
  \label{fig:motivation}
\end{figure}

However, both the \textit{sim/real domain gap} and the \textit{inter-agent domain gap} hinder the generalization of the model trained on simulated datasets to real-world applications. 
Simulated data and real-world data differ in many aspects, \eg ~sensor type, reflectance patterns, and road surroundings. 
\Cref{fig:motivation} (a) compares the point clouds of ego agents from simulated V2XSet \cite{v2xvit} and real-world DAIR-V2X-C \cite{dair} dataset.
As \Cref{fig:motivation} (a) shows, the significant sim/real domain gap results in point clouds that differ in density, distribution, reflectance, \etc , posing a challenge to the model's generalization ability.
Moreover, in real-world applications, collaborative agents can often be heterogeneous, equipped with multiple types of sensors with various extrinsics. 
The heterogeneity of agents can lead to a prominent inter-agent domain gap.
\Cref{fig:motivation} (b) shows the severe domain gap between a data acquisition vehicle and a roadside infrastructure in the DAIR-V2X-C \cite{dair} dataset.
A $40$-beam mechanical LiDAR is installed on the top of the vehicle, whereas a $300$-beam solid-state LiDAR is placed on the roadside pole.
The divergence in sensor type and placement also results in distinct point clouds, further increasing the difficulty of sim2real adaptation. 
These two domain gaps mentioned above usually lead to a serious performance drop when a model is trained on simulated collaborative perception datasets but evaluated on real-world collaborative perception datasets.

To take full advantage of simulated data and appropriately address the aforementioned domain gaps, we propose a novel Decoupled Unsupervised Sim2Real Adaptation (DUSA) method for V2X collaborative detection.
Our new method decouples the V2X collaborative sim2real domain adaptation problem into two sub-problems: sim2real adaptation and inter-agent adaptation and then conquers them respectively. 
%
For sim2real adaptation, we design a Location-adaptive Sim2Real Adapter (LSA) module to align the feature maps of ego agents between the simulation and the real world. 
It first leverages a Location-adaptive Feature Selector to capture the internal pattern of significance distribution and aggregate features of great significance from the ego agent's feature map accordingly.
It produces a global feature for each ego agent respectively. 
Then, it aligns the aggregated global feature via a sim/real discriminator.
For inter-agent adaptation, we further introduce a Confidence-aware Inter-agent Adapter (CIA) module to align the features from heterogeneous real-world agents, \eg ~ data acquisition vehicles and roadside infrastructures.
It discriminates the fine-grained feature under the guidance of agent-wise confidence maps.
%
The two decoupled adapters encourage the model to output both sim/real-invariant and agent-invariant feature maps, boosting the generalization capability of the adapted model.
Experiments demonstrate the effectiveness of our design and its superiority over naive sim/real discriminator and self-training baseline. 

In summary, our main contributions can be listed as follow:

\begin{itemize}
    

    \item We introduce DUSA, a new unsupervised sim2real domain adaptation method for V2X collaborative detection, which decouples the V2X sim2real domain adaptation problem into two sub-problems: sim2real adaptation and inter-agent adaptation and addresses them respectively.
    
    \item We propose to utilize location significance and confidence clues to facilitate the adapters, and devise the LSA module and the CIA module. The two modules bridge the gap between sim/real and heterogeneous agents, respectively.

    \item Extensive experiments demonstrate the effectiveness of the proposed DUSA approach for V2X unsupervised sim2real adaptation on several collaborative 3D detectors from the V2XSet dataset \cite{v2xvit} to the DAIR-V2X-C dataset \cite{dair}. The code is available at \href{https://github.com/refkxh/DUSA}{https://github.com/refkxh/DUSA}.
\end{itemize}

\section{Related Work}

\subsection{Collaborative 3D Object Detection}
Utilizing cooperative perception to support autonomous driving of single vehicles has become a topic of increasing research interest \cite{hobert2015enhancements, liu2021towards}. 
Method \cite{arnold2020cooperative} first introduces early and late fusion schemes in a cooperative perception system between vehicles and infrastructure. The WIBAM model \cite{WIBAM} proposes a technique for fine-tuning traffic observation camera models using weak supervision. Cooper \cite{cooper} utilizes raw point clouds from multiple vehicles and develops the SPOD network. F-Cooper \cite{fcooper} realizes feature-level fusion based on cooper, which reduces the amount of communication data while ensuring accuracy. The V2VNet \cite{v2vnet} introduces the intermediate fusion method, which applies a graph neural network that is aware of spatial relationships to merge data from multiple vehicles. DiscoNet \cite{disconet} employs knowledge distillation to train a DiscoGraph, which helps to achieve a superior balance between performance and bandwidth usage in multi-agent perception. V2X-ViT \cite{v2xvit} proposes a vision transformer framework to achieve feature fusion between vehicles and infrastructure. SyncNet \cite{lei2022latency} solves the time-domain synchronization problem. Where2comm \cite{Where2comm:22} proposes the use of spatial confidence maps to reduce the amount of communication bandwidth required. This is achieved by limiting the transfer of unnecessary data. CoBEVT \cite{xu2022cobevt} utilizes fused axial attention to cooperatively generate bird's eye view (BEV) predictions across multiple agents and cameras. AdaFusion \cite{Qiao_2023_WACV} proposes three adaptive models for feature fusion in the bird's eye view (BEV) to enhance perception accuracy. MPDA \cite{xu2023mpda} identifies a domain gap issue that arises between different agents and addresses it by utilizing complete feature maps to standardize patterns. However, this method requires significant communication resources. Method \cite{modelagnostic} proposes a model-agnostic perception system to solve the problem of mismatching in multi-agent confidence distributions. 
The proposed DUSA is applicable to multiple collaborative 3D detectors, \eg ~V2X-ViT \cite{v2xvit}, DiscoNet \cite{disconet}, and F-Cooper \cite{fcooper}.

To advance the progress of cooperative perception, numerous high-quality datasets have been released publicly in recent years. 
OPV2V \cite{opv2v}, V2X-Sim \cite{v2xsim}, and V2XSet \cite{v2xvit} are simulated datasets that specifically address the issue of multi-agent cooperative perception. 
DAIR-V2X-C \cite{dair} is the only real-world vehicle-infrastructure cooperative benchmark that supports both camera and LiDAR modalities. However, originally DAIR-V2X-C only annotates 3D boxes within the range of the vehicle-side camera's view. To enable 360-degree detection, CoAlign \cite{coalign} supplements the missing 3D box annotations. 
We select the simulated V2XSet \cite{v2xvit} dataset as our annotated source domain and the real-world DAIR-V2X-C \cite{dair} dataset as the unlabeled target domain.

\subsection{Unsupervised Domain Adaptation}

Unsupervised Domain Adaptation (UDA) aims to produce a robust model that can generalize effectively to target domains with labeled source data and unlabeled target data. Many works \cite{ganin2015unsupervised, hoffman2016fcns, ganin2016domain, tsai2018learning,chen2018domain, cdal, interbn, ic2fa, imse, m2guda, emotiongan, sdfuda} utilize adversarial learning \cite{NIPS2014_5ca3e9b1} to align feature distributions across different domains by minimizing the H-divergence \cite{ben2010theory} or the Jensen-Shannon divergence \cite{gulrajani2017improved} between two domains. Other kinds of methods \cite{khodabandeh2019robust, zou2018unsupervised, saito2017asymmetric, sdfuda} develop a variety of pseudo labels for unlabeled target domains to achieve self-training \cite{lee2013pseudo}. 

As for the domain adaptation for 3D object detection, 
PointDAN \cite{qin2019pointdan} is the first method to employ adversarial learning to match point cloud distributions between domains with non-autonomous driving scenarios. SN \cite{wang2020train} normalizes object sizes in the source domain by utilizing object statistics from the target domain to reduce the domain gap at the size level. 
SRDAN \cite{zhang2021srdan} proposes scale-aware and range-aware domain alignment strategies that leverage the geometric characteristics of 3D data to guide the distribution alignment between two domains. MLC-Net \cite{luo2021unsupervised} employs a teacher-student paradigm and exploits multi-level consistency to facilitate cross-domain transfer. ST3D \cite{Yang_2021_CVPR} proposes a comprehensive pipeline for pseudo-label generation and training process based on the random object scaling strategy. Moreover, ST3D++ \cite{yang2022st3d++} further proposes an optimization strategy for denoising pseudo-labels in the context of 3D UDA. 
Our DUSA is an adversarial-learning-based UDA method without the need for the target domain's annotations.

\section{Method}

\subsection{Problem Statement}
In this work, we address the unsupervised sim2real domain adaptation problem for V2X collaborative 3D detection. 
In this problem, we have a set of simulated samples with labels as the source domain and a set of unlabeled real-world samples as the target domain. 
Each sample contains multiple collaborative agents, and each agent includes a LiDAR point cloud and their poses under the world coordinate. 
Collaborative agents could be heterogeneous in sensor type and placement, \etc~
We expect the model could fully exploit the annotated simulated samples to improve its performance on the unlabeled real-world samples.

Formally, let us denote a point cloud captured by a collaborative agent as $\boldsymbol{P}_j=\{\boldsymbol{p}_k\mid _{k=1}^{n_k}\}$, where $\boldsymbol{p}_k$ is a point containing 3D coordinates and other features (\eg ~intensity), $n_k$ is the number of points in $\boldsymbol{P}_j$, and $j$ is the agent number. 
We denote the agent pose under the world coordinate as $\boldsymbol{T}_j \in SE(3)$, where $j$ is the agent number. 
We further denote a collaborative sample as $\boldsymbol{X}=\{(\boldsymbol{P}_j, \boldsymbol{T}_j)\mid _{j=1}^{n_a}\}$, where $n_a$ is the number of agents in $\boldsymbol{X}$.
For the labeled source domain, we denote it as $\mathcal{X}^s=\{(\boldsymbol{X}_i^s, \mathcal{Y}_i^s)\mid _{i=1}^{N_s}\}$, where $\mathcal{Y}_i^s$ is the corresponding annotation, and $N_s$ is the total number of source domain samples.
Similarly, we have $\mathcal{X}^t=\{\boldsymbol{X}_i^t\mid _{i=1}^{N_t}\}$ as the unlabeled target domain, where $N_t$ is the number of target domain samples. 
Our goal is to train a robust detector which can accurately predict the bounding boxes for all samples in the unlabeled target domain.

\subsection{Framework}

\begin{figure*}[t]
  \centering
  \includegraphics[width=0.95\textwidth]{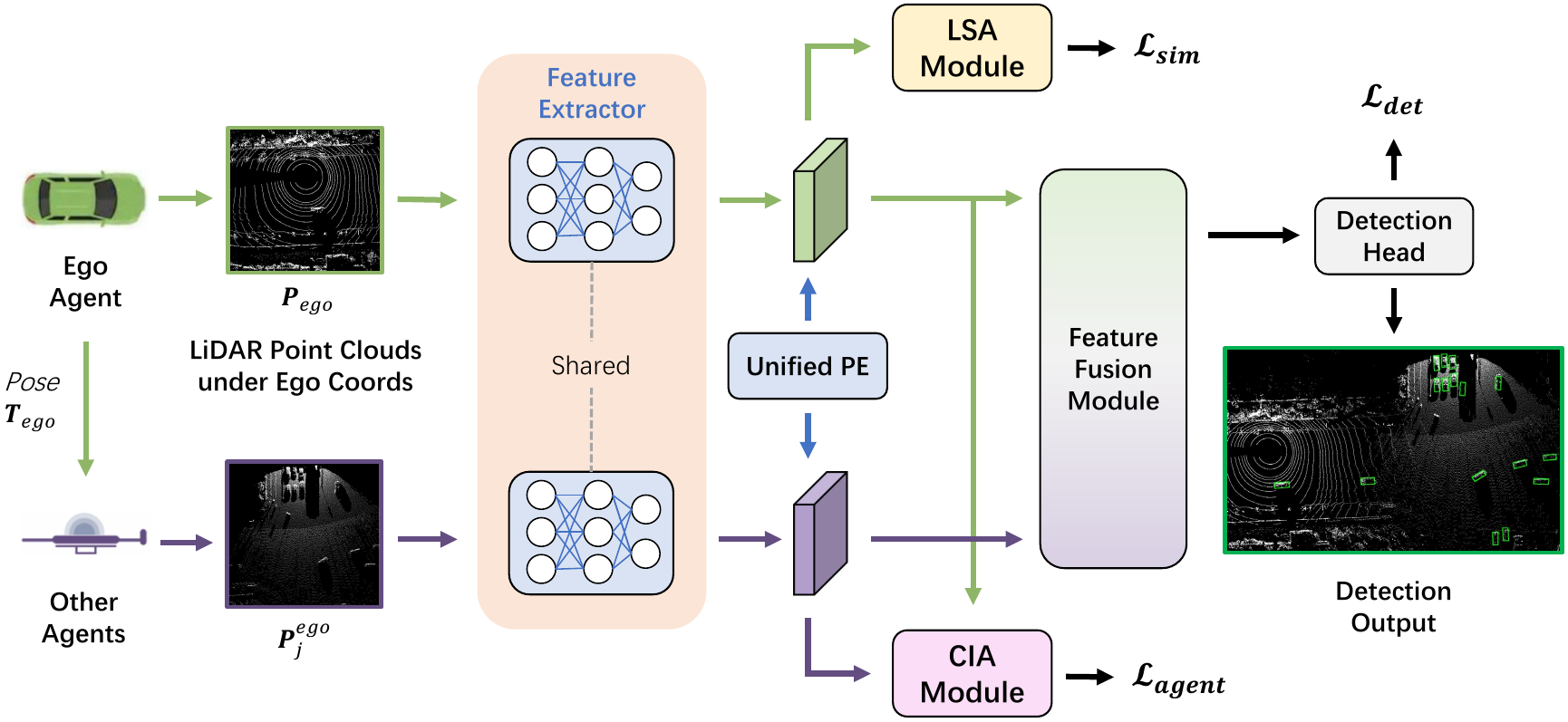}
  \caption{Framework of the proposed DUSA method. The model first extracts features from agent-wise LiDAR point clouds under ego coordinates and then fuses the agent-wise features to predict the collaborative detection results. The sim/real and inter-agent domain gaps are handled by the decoupled LSA module and CIA module, respectively. The two modules encourage the feature extractor to produce sim/real-invariant and agent-invariant features through adversarial training.}
  \Description{Framework of DUSA.}
  \label{fig:method}
\end{figure*}

To achieve our goal, we introduce DUSA to bridge the sim/real and inter-agent domain gaps respectively. 
The proposed DUSA can be applied to most of the typical collaborative 3D detectors based on intermediate fusion. 
\Cref{fig:method} shows the framework of DUSA.

We first introduce the standard process of intermediate-fusion collaborative 3D detectors.
In the beginning, the ego agent sends its pose $\boldsymbol{T}_{ego}$ to other collaborative agents.
Then, other agents project their point clouds $\boldsymbol{P}_j$ to ego agent's coordinates following \Cref{eq:projection}.

\begin{equation}
    \boldsymbol{P}_j^{ego}=\boldsymbol{T}_{ego}^{-1}\boldsymbol{T}_j\boldsymbol{P}_j
    \label{eq:projection}
\end{equation}

\noindent After projection, a point cloud feature extractor is utilized to generate feature maps for each agent. 
We denote the extracted feature as $F(\boldsymbol{P}_j^{ego};\boldsymbol{\theta})$, where $\boldsymbol{\theta}$ stands for the feature extractor's parameters shared among all agents. 
For conciseness, we denote the extracted features of all agents in the i-th collaborative sample as $\boldsymbol{F}(\boldsymbol{X}_i;\boldsymbol{\theta})$.
Next, a feature fusion module integrates features from all agents.
The integrated feature is represented as $G(\boldsymbol{F}(\boldsymbol{X}_i;\boldsymbol{\theta}); \boldsymbol{\phi})$, where $\boldsymbol{\phi}$ is the parameters of the feature fusion module. 
Finally, a detection head predicts the final results using the integrated feature, which can be denoted as $P(G(\boldsymbol{F}(\boldsymbol{X}_i;\boldsymbol{\theta}); \boldsymbol{\phi}); \boldsymbol{\beta})$, where $\boldsymbol{\beta}$ is the parameters of the detection head.

Before adaptation, a unified positional encoding is concatenated to the features from every agent. 
We denote the features with positional encoding as $\tilde{\boldsymbol{F}}(\boldsymbol{X}_i;\boldsymbol{\theta})$. 
The unified positional encoding is a 2D coordinate representing the relative distance to the ego agent. 
In this way, the locations with similar distances to the ego agent will be assigned the same coordinate. 

The LSA module and the CIA module align features from different domains and agents in the training process. 
The LSA module takes in the feature of ego agent and discriminates whether the agent is from simulation or the real world. 
The CIA module takes in the features and confidence maps of each real-world agent and discriminates the type of agent. 
The two modules encourage the feature extractor to produce sim/real-invariant and agent-invariant features through adversarial training, where the training objectives of the two modules and the feature extractor are opposite.
We detail the two modules in \Cref{sec:lsa,sec:cia}.

Compared to simply discriminating whether the agent is simulated or real for all collaborative agents, DUSA decouples the discrimination problem into two more specific and less entangled sub-tasks, \ie ~sim/real discrimination and agent-type discrimination. 
The decoupled tasks allow the discriminators to find more detailed domain gaps, and increase the domain invariance of the feature extractor through adversarial training.

\subsection{Location-adaptive Sim2Real Adapter}
\label{sec:lsa}

The Location-adaptive Sim2Real Adapter (LSA) module aims to bridge the domain gap between simulated and real-world data. 
To fully exclude the inter-agent domain gap from the sim/real domain gap, the LSA module only discriminates features from agents equipped with relatively similar sensors. 
We select the ego agents' feature $\tilde{F}(\boldsymbol{P}_{ego};\boldsymbol{\theta})$ from both domains to discriminate, for they usually have a mechanical LiDAR installed on the top no matter they are from simulation or real world.

\begin{figure}[t]
  \centering
  \includegraphics[width=\linewidth]{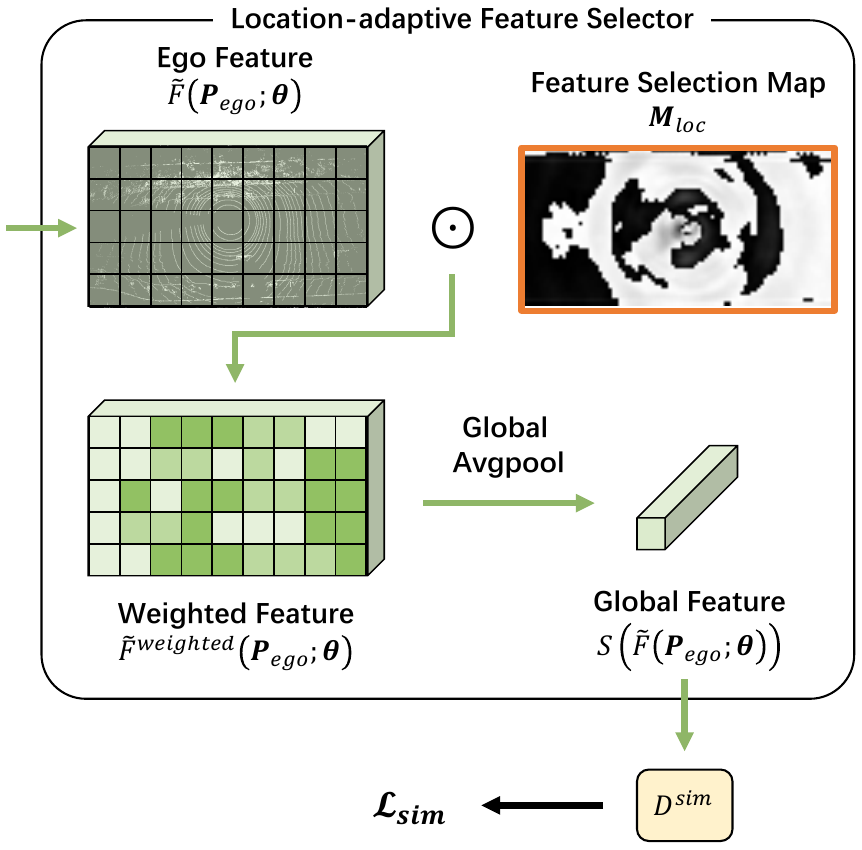}
  \caption{Architecture of the LSA module. It only discriminates features from ego agents of both domains to bridge the sim/real domain gap. The Location-adaptive Feature Selector captures the internal pattern of the grid's significance distribution and aggregates a global feature from the input feature map accordingly. Then a sim/real discriminator is used for adversarial training.}
  \Description{Details of the LSA module.}
  \label{fig:lsa}
\end{figure}

\Cref{fig:lsa} displays the internal architecture of the LSA module. 
For an autonomous vehicle, different locations on its feature map often have various significance for the discrimination task. 
The distribution of significant locations for a single agent has its internal pattern. 
For instance, grids that are near the agent often possess more domain clues than grids that are far from the agent since they usually have more LiDAR points on nearby vehicles. 
Moreover, grids that are in front the vehicle are usually more significant than grids that are behind the vehicle. 
Thus, we devise a Location-adaptive Feature Selector module to adaptively select features of great importance according to the distribution. 
Specifically, we use a learnable feature selection map $\boldsymbol{M}_{loc}$ to capture the internal pattern of the distribution and select features of great significance from the ego agent's feature map accordingly. 
Firstly, we reweigh the input feature map $\tilde{F}(\boldsymbol{P}_{ego};\boldsymbol{\theta})$ by conducting element-wise multiplication between the feature map and the learnable feature selection map $\boldsymbol{M}_{loc}$. 
The weighted feature map is denoted as \Cref{eq:weighting}.

\begin{equation}
    \tilde{F}^{weighted}(\boldsymbol{P}_{ego};\boldsymbol{\theta})=\tilde{F}(\boldsymbol{P}_{ego};\boldsymbol{\theta})\odot\boldsymbol{M}_{loc}
    \label{eq:weighting}
\end{equation}

\noindent Then, we use the global average pooling operation to obtain a global feature as \Cref{eq:avgpool},

\begin{equation}
    S(\tilde{F}(\boldsymbol{P}_{ego};\boldsymbol{\theta}))=\frac{1}{HW}\sum_{u,v}\tilde{F}^{weighted}(\boldsymbol{P}_{ego};\boldsymbol{\theta})^{(u,v)}
    \label{eq:avgpool}
\end{equation}

\noindent where $H$ and $W$ denote the height and width of the input feature.

We perform sim/real discrimination on the aggregated global feature $S(\tilde{F}(\boldsymbol{P}_{ego};\boldsymbol{\theta}))$. 
Similar to \cite{grl}, our sim/real discriminator includes several fully connected layers with ReLU activation and dropout. 
We denote it as $D^{sim}(\cdot;\boldsymbol{w}^{sim})$, where $\boldsymbol{w}^{sim}$ are the parameters of the sim/real discriminator. 
The objective of sim2real adaptation can be written as \Cref{eq:sim2real},

\begin{equation}
    \mathop{max}\limits_{\boldsymbol{\theta}}\mathop{min}\limits_{\boldsymbol{w}^{sim}}\mathcal{L}_{sim}=\frac{1}{N}\sum_{i=1}^N\mathcal{L}_{BCE}(D^{sim}(S(\tilde{F}(\boldsymbol{P}_{ego};\boldsymbol{\theta}));\boldsymbol{w}^{sim}),d_i)
    \label{eq:sim2real}
\end{equation}

\noindent where $N=N_s+N_t$ is the total number of source and target samples, $\mathcal{L}_{BCE}$ is the binary cross-entropy loss, and $d_i\in\{0,1\}$ indicates whether the training sample comes from the source (simulated) domain or the target (real-world) domain.

To jointly optimize the max-min optimization problem in \Cref{eq:sim2real}, we insert a gradient reversal layer (GRL) \cite{grl} before the input of the LSA module. By doing so, the optimization can be done in a single step.

\subsection{Confidence-aware Inter-agent Adapter}
\label{sec:cia}

The Confidence-aware Inter-agent Adapter (CIA) module aims to minimize the domain gap between heterogeneous agents in the real world. 
It focuses on discriminating the real-world agents' fine-grained feature maps $\tilde{\boldsymbol{F}}(\boldsymbol{X}_i^t;\boldsymbol{\theta})$, excluding the domain gap between simulation and real world.

\begin{figure}[t]
  \centering
  \includegraphics[width=0.7\linewidth]{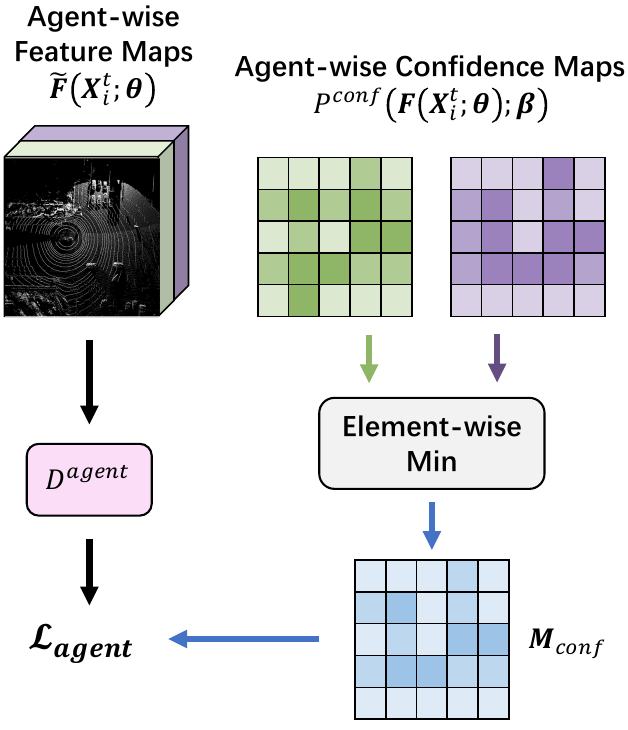}
  \caption{Details of the CIA module. It focuses on discriminating the heterogeneous real-world agents' fine-grained feature maps to minimize the inter-agent domain gap. The inter-agent discriminator loss is reweighed under the guidance of agent-wise confidence maps to avoid discriminating \textit{almost-void} grids and stabilize the optimization.}
  \label{fig:cia}
\end{figure}

We detail the proposed CIA module in \Cref{fig:cia}. 
In this module, we perform inter-agent discrimination on agent-wise fine-grained feature maps $\tilde{\boldsymbol{F}}(\boldsymbol{X}_i^t;\boldsymbol{\theta})$ from the target domain (real world). 
The inter-agent discriminator includes several $1\times 1$ convolutional layers with ReLU activation. 
We denote it as $D^{agent}(\cdot;\boldsymbol{w}^{agent})$, where $\boldsymbol{w}^{agent}$ is the parameters of the inter-agent discriminator. 

In agent-wise fine-grained feature maps, there are some \textit{almost-void} grids that have very few points in them. 
Consequently, these grids include very few clues about the sensors and are inappropriate to perform inter-agent discrimination. 
Supervising the inter-agent discriminator on these grids may include noise in the gradients and increase the instability of the optimization process. 
To address this issue, we propose to utilize agent-wise confidence maps to reweigh the inter-agent discrimination loss. 
Grids with high confidence scores among all agents are very likely to be \textit{non-void}. 
Specifically, we first obtain the agent-wise confidence maps $P^{conf}(\boldsymbol{F}(\boldsymbol{X}_i^t;\boldsymbol{\theta}); \boldsymbol{\beta})$ with the help of the detection head, and then perform element-wise minimization among each agent to get the reweighing map $\boldsymbol{M}_{conf}$ as \Cref{eq:conf-map},

\begin{equation}
    \boldsymbol{M}_{conf}^{(u,v)}=\mathop{min}\limits_{j}P^{conf}(\boldsymbol{F}(\boldsymbol{X}_i^t;\boldsymbol{\theta})_j; \boldsymbol{\beta})^{(u,v)}
\label{eq:conf-map}
\end{equation}

\noindent where $j$ stands for the agent number. The reweighing map $\boldsymbol{M}_{conf}$ is then used for re-balancing the loss of each grid.

The objective of inter-agent adaptation can be written as \Cref{eq:inter-agent},

\begin{equation}
\begin{split}
    \mathop{max}\limits_{\boldsymbol{\theta}}\mathop{min}\limits_{\boldsymbol{w}^{agent}}\mathcal{L}_{agent}=&\frac{1}{N_t\cdot n_a}\sum_{i=1}^{N_t}\sum_{j=1}^{n_a}\sum_{u,v}\boldsymbol{M}_{conf}^{(u,v)}\cdot\\
    &\mathcal{L}_{CE}(D^{agent}(\tilde{\boldsymbol{F}}(\boldsymbol{X}_i^t;\boldsymbol{\theta})_j;\boldsymbol{w}^{agent})^{(u,v)},j)
\end{split}
\label{eq:inter-agent}
\end{equation}

\noindent where $\mathcal{L}_{CE}$ is the cross-entropy loss, and $j$ indicates the agent number. A gradient reversal layer (GRL) \cite{grl} is also inserted before the input of the CIA module, \ie ~real-world agents' fine-grained feature maps $\tilde{\boldsymbol{F}}(\boldsymbol{X}_i^t;\boldsymbol{\theta})$, for optimizing this objective function.

\subsection{Overall Training Objectives}

The overall training objectives of DUSA consist of three components: the detection loss $\mathcal{L}_{det}$, the sim/real discrimination loss $\mathcal{L}_{sim}$, and the inter-agent discrimination loss $\mathcal{L}_{agent}$.

For the detection loss, we use similar loss functions as PointPillars \cite{pointpillars} except for the softmax classification loss on the discretized directions $\mathcal{L}_{dir}$. 
The detection loss includes a localization loss $\mathcal{L}_{loc}$, which is a Smooth-L1 loss, and an object classification loss $\mathcal{L}_{cls}$, which is a focal loss \cite{focalloss}. 
We denote the sum of the two loss functions as $\mathcal{L}_{PP}$. 
Since we only have annotations in the source domain $\mathcal{X}^s$, the detection loss is only applied to the source domain. 
Formally, we denote the detection loss as \Cref{eq:ldet}. 

\begin{equation}
    \mathcal{L}_{det}=\frac{1}{N_s}\sum_{i=1}^{N_s}\mathcal{L}_{PP}(P(G(\boldsymbol{F}(\boldsymbol{X}_i;\boldsymbol{\theta}); \boldsymbol{\phi}); \boldsymbol{\beta}),\mathcal{Y}_i^s)
    \label{eq:ldet}
\end{equation}

\noindent For more details about the detection loss, please refer to \cite{pointpillars}. 

Taking all the objectives into account, we rewrite the combination of \Cref{eq:sim2real,eq:inter-agent,eq:ldet} as \Cref{eq:overall} as the overall objective,

\begin{equation}
    \mathop{min}\limits_{\boldsymbol{\theta}}(\mathop{min}\limits_{\boldsymbol{\phi ,\beta}}\mathcal{L}_{det}-\mathop{min}\limits_{\boldsymbol{w}^{sim}}\alpha _1\mathcal{L}_{sim}-\mathop{min}\limits_{\boldsymbol{w}^{agent}}\alpha _2\mathcal{L}_{agent})
    \label{eq:overall}
\end{equation}

\noindent where $\alpha _1$ and $\alpha _2$ are scalars used for balancing the loss terms. 
We optimize this goal through adversarial training with GRLs.
 
\section{Experiments}

\subsection{Datasets}

To validate the effectiveness of the proposed DUSA method, we select a simulated V2X collaborative detection dataset V2XSet \cite{v2xvit} as the annotated source domain $\mathcal{X}^s$ and a real-world V2X dataset DAIR-V2X-C \cite{dair} as the unlabeled target domain $\mathcal{X}^t$. 
We conduct unsupervised domain adaptation on the two datasets following the standard protocol.

\subsubsection{V2XSet}
V2XSet \cite{v2xvit} is a large-scale simulated V2X 3D detection dataset based on CARLA \cite{carla} and OpenCDA \cite{opencda}. 
It has 11,447 frames in total, 6,694 for training, 1,920 for validation, and 2,833 for testing. 
In each scene, there are 2-7 agents for collaborative perception. 
Note that all the vehicles are \textit{cars} in V2XSet.

\subsubsection{DAIR-V2X-C}
DAIR-V2X-C \cite{dair} is the first large-scale, real-world dataset for vehicle-infrastructure cooperative detection. 
It contains LiDAR point clouds and extrinsics from both vehicles and infrastructures. 
DAIR-V2X-C has 38,845 camera frames and 38,845 Lidar frames in total. 
Since there are no other types of vehicles except for \textit{cars} in V2XSet, we discard all the other types of vehicles in DAIR-V2X-C and only focus on \textit{cars} during the evaluation process. 
Since DAIR-V2X-C only annotates 3D boxes within the range of the vehicle-side camera's view, we use the supplemented annotation in CoAlign \cite{coalign} in the evaluation process for $360^{\circ}$ detection.

\subsection{Implementation Details}
We use the OpenCOOD \cite{opv2v} toolbox to implement the proposed DUSA method. 
We select PointPillars \cite{pointpillars} as the point cloud feature extractor. 
The input data range is set to $[-102.4m,102,4m]$ along the x-axis,  $[-38.4m,38.4m]$ along the y-axis, and $[-3.5m,1.5m]$ along the z-axis. 
We assume that there are no localization noise, asynchronization, and transmission latency between collaborative agents, \ie ~the \textit{perfect} setting.

We first train the collaborative detector on the source domain only without the domain adapters until converged, and then apply the LSA and the CIA on the two domains jointly. 
The model is optimized by the Adam optimizer with a batch size of $4$ ($2$ from each domain). 
The initial learning rate is set to $0.001$ and is decayed by the factor of $0.1$ after $15$ epochs. 
The scalars $\alpha _1$ and $\alpha _2$ in the overall training objective are both set to $1$ empirically. 
The gradient reversal factors in GRLs are set to $-0.05$ for the LSA module and $-0.1$ for the CIA module, respectively. 
The data augmentation in OpenCOOD \cite{opv2v} is applied in the adaptation process.
Other settings follow the default config in OpenCOOD \cite{opv2v}. 
All experiments are implemented with PyTorch on a single NVIDIA V100 GPU.

\subsection{Baselines}

To the best of our knowledge, there is still no method addressing the unsupervised sim2real adaptation problem specially for V2X collaborative detection. 
To better evaluate the effectiveness of the proposed DUSA method, we implemented two simple baselines for this problem.

\subsubsection{Self-training}
The self-training baseline alternates between pseudo-label generation and finetuning. 
Specifically, the collaborative detection model is first trained on the labeled source domain until converged and then predicts bounding boxes with confidence scores for the unlabeled target domain. 
A confidence threshold is applied to the predictions to obtain pseudo labels. 
Then, the model finetunes itself on the target domain with generated pseudo labels. 
The baseline method alternates between the two progresses and iteratively learns the new distribution of the target domain.

\subsubsection{Naive Sim/real Discriminator}
The naive sim/real discriminator simply discriminates whether the collaborative agent is from simulation or the real world, regardless of the heterogeneity among agents. 
In detail, we also pretrain the collaborative detection model on the annotated source domain until converged. 
Then, a discriminator similar to $D^{sim}$ is applied to distinguish whether the extracted features of all agents $\boldsymbol{F}(\boldsymbol{X}_i;\boldsymbol{\theta})$ is from simulation or real world. 
The discriminator is optimized jointly with the detector through a GRL. 
Supervision for the naive sim/real discriminator is also similar to that for the LSA module.

\subsection{Quantitative Evaluation}

To demonstrate the effectiveness and generalizability of the proposed DUSA method, we apply DUSA to three commonly-used collaborative 3D detection methods: V2X-ViT \cite{v2xvit}, DiscoNet \cite{disconet} (student only), and F-Cooper \cite{fcooper} and conduct a quantitative evaluation. 
The evaluation is performed on the validation set of the DAIR-V2X-C dataset. 
The evaluation metric is AP @ $m$, which means the average precision with IoU threshold $m$, where $m\in\{0.3,0.5,0.7\}$.

\subsubsection{V2X-ViT}

\begin{table}
  \caption{Quantitative evaluation on V2X-ViT. 
  \textcolor{gray}{Gray texts} stand for the AP gain compared to \textit{No adaptation}. DUSA outperforms the baseline methods by a large margin on V2X-ViT.}
  \label{tab:v2xvit}
  \begin{tabular}{lccc}
    \toprule
    Method & AP @ 0.3 & AP @ 0.5 & AP @ 0.7 \\
    \midrule
    Oracle & 62.69 & 59.02 & 45.61 \\
    No adaptation & 34.73 & 31.23 & 14.94 \\
    \midrule
    Self-training & 33.03 \textcolor{gray}{(-1.70)} & 31.67 \textcolor{gray}{(+0.44)} & 19.01 \textcolor{gray}{(+4.07)} \\
    Discriminator & 38.54 \textcolor{gray}{(+3.81)} & 33.89 \textcolor{gray}{(+2.66)} & 15.13 \textcolor{gray}{(+0.19)} \\
    \textbf{DUSA (Ours)} & \textbf{43.61 \textcolor{gray}{(+8.88)}} & \textbf{38.46 \textcolor{gray}{(+7.23)}} & \textbf{20.09 \textcolor{gray}{(+5.15)}} \\
  \bottomrule
\end{tabular}
\end{table}

The quantitative results of different domain adaptation methods on V2X-ViT are summarized in \Cref{tab:v2xvit}. 
\textit{Oracle} means the V2X-ViT model is both trained and evaluated on the target domain, \ie ~the DAIR-V2X-C dataset, and there is no domain gap between training and evaluation. 
\textit{No adaptation} means the V2X-ViT model trained on the source domain, \ie ~the V2XSet dataset, is directly evaluated on the target domain, \ie ~the DAIR-V2X-C dataset without any domain adaptation techniques. 

As shown in \Cref{tab:v2xvit}, DUSA outperforms the baseline methods by a large margin on V2X-ViT. 
Compared to the self-training baseline, DUSA has improved by $+10.58\%$ at AP @ 0.3, $+6.79\%$ at AP @ 0.5, and $+1.08\%$ at AP @ 0.7. 
As for the naive sim/real discriminator, DUSA also outperforms it by $+5.07\%$ at AP @ 0.3, $+4.57\%$ at AP @ 0.5, and $+4.96\%$ at AP @ 0.7.

\subsubsection{F-Cooper}

\begin{table}
  \caption{Quantitative evaluation on F-Cooper. \textcolor{gray}{Gray texts} stand for the AP gain compared to \textit{No adaptation}. DUSA significantly surpasses the baseline methods on F-Cooper.}
  \label{tab:fcooper}
  \begin{tabular}{lccc}
    \toprule
    Method & AP @ 0.3 & AP @ 0.5 & AP @ 0.7 \\
    \midrule
    Oracle & 61.26 & 57.19 & 45.04 \\
    No adaptation & 35.12 & 30.27 & 13.46 \\
    \midrule
    Self-training & 40.46 \textcolor{gray}{(+5.34)} & 35.89 \textcolor{gray}{(+5.62)} & 16.38 \textcolor{gray}{(+2.92)} \\
    Discriminator & 41.33 \textcolor{gray}{(+6.24)} & 36.20 \textcolor{gray}{(+5.93)} & 16.81 \textcolor{gray}{(+3.35)} \\
    \textbf{DUSA (Ours)} & \textbf{42.83 \textcolor{gray}{(+7.71)}} & \textbf{37.93 \textcolor{gray}{(+7.66)}} & \textbf{20.25 \textcolor{gray}{(+6.79)}} \\
  \bottomrule
\end{tabular}
\end{table}

Quantitative results in \Cref{tab:fcooper} display the effectiveness of different domain adaptation methods on F-Cooper. 
DUSA significantly surpasses the baseline methods on F-Cooper. 
Compared to the self-training baseline, DUSA has improved by $+2.37\%$ at AP @ 0.3, $+2.04\%$ at AP @ 0.5, and $+3.87\%$ at AP @ 0.7. 
As for the naive sim/real discriminator, DUSA also surpasses it by $+1.50\%$ at AP @ 0.3, $+1.73\%$ at AP @ 0.5, and $+3.44\%$ at AP @ 0.7. 
AP @ 0.7 increases the most on F-Cooper, which may indicate that DUSA is capable to improve the localization accuracy.

\subsubsection{DiscoNet}

\begin{table}
  \caption{Quantitative evaluation on DiscoNet (student only). \textcolor{gray}{Gray texts} stand for the AP gain compared to \textit{No adaptation}. DUSA achieves prominent improvement on DiscoNet.}
  \label{tab:disconet}
  \begin{tabular}{lccc}
    \toprule
    Method & AP @ 0.3 & AP @ 0.5 & AP @ 0.7 \\
    \midrule
    Oracle & 62.87 & 58.53 & 46.69 \\
    No adaptation & 35.11 & 31.07 & 16.93 \\
    \midrule
    Self-training & 30.66 \textcolor{gray}{(-4.45)} & 28.72 \textcolor{gray}{(-2.35)} & 17.26 \textcolor{gray}{(+0.33)} \\
    Discriminator & 39.55 \textcolor{gray}{(+4.44)} & 34.37 \textcolor{gray}{(+3.30)} & 19.05 \textcolor{gray}{(+2.12)} \\
    \textbf{DUSA (Ours)} & \textbf{40.03 \textcolor{gray}{(+4.92)}} & \textbf{35.23 \textcolor{gray}{(+4.16)}} & \textbf{20.68 \textcolor{gray}{(+3.75)}} \\
  \bottomrule
\end{tabular}
\end{table}

We implemented a student-only DiscoNet model to further validate the generalizability of DUSA.
As displayed in \Cref{tab:disconet}, DUSA achieves prominent improvement on DiscoNet.
Compared to the self-training baseline, DUSA has improved by $+9.37\%$ at AP @ 0.3, $+6.51\%$ at AP @ 0.5, and $+3.42\%$ at AP @ 0.7. 
As for the naive sim/real discriminator, DUSA also exceeds it by $+0.48\%$ at AP @ 0.3, $+0.86\%$ at AP @ 0.5, and $+1.63\%$ at AP @ 0.7. 

\subsection{Ablation Studies}

\begin{figure*}[ht]
  \centering
  \includegraphics[width=0.84\textwidth]{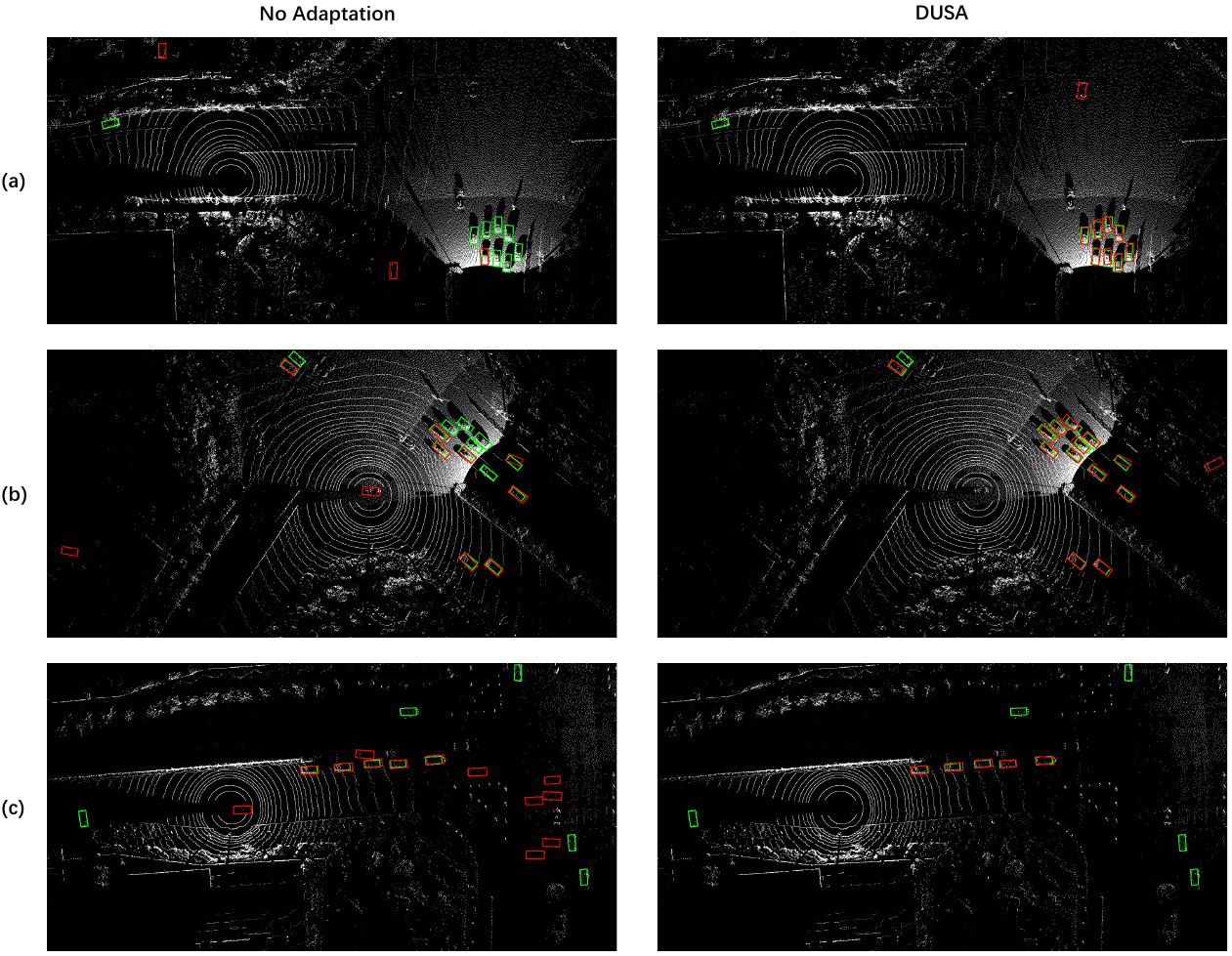}
  \caption{Visualization of detection results of the V2X-ViT model without domain adaptation and the DUSA-adapted V2X-ViT model. \textcolor{green}{Green boxes} are ground truths, and \textcolor{red}{red boxes} are predictions.}
  \Description{Qualitative results of DUSA.}
  \label{fig:vis}
\end{figure*}

To validate the effectiveness of our proposed decoupled sim/real and inter-agent domain alignment strategies, we further conduct comprehensive ablation studies by varying different components in our DUSA. 
The ablations are done on the validation split of the DAIR-V2X-C \cite{dair} dataset with the V2X-ViT \cite{v2xvit} detector.

\begin{table}
  \caption{Ablation studies of DUSA on V2X-ViT. \textcolor{gray}{Gray texts} stand for the AP gain compared to \textit{No adaptation}.}
  \label{tab:ablations}
  \begin{tabular}{lccc}
    \toprule
    Method & AP @ 0.3  & AP @ 0.5 & AP @ 0.7 \\
    \midrule
    Oracle & 62.69 & 59.02 & 45.61 \\
    No adaptation & 34.73 & 31.23 & 14.94 \\
    \midrule
    + LSA w/o LFS & 38.87 \textcolor{gray}{(+4.14)} & 34.61 \textcolor{gray}{(+3.38)} & 15.76 \textcolor{gray}{(+0.82)} \\
    + LSA & 39.21 \textcolor{gray}{(+4.48)} & 34.83 \textcolor{gray}{(+3.60)} & 17.1 \textcolor{gray}{(+2.16)} \\
    \midrule
    + CIA w/o Conf & 39.28 \textcolor{gray}{(+4.55)} & 34.63 \textcolor{gray}{(+3.40)} & 16.14 \textcolor{gray}{(+1.20)} \\
    + CIA & 39.06 \textcolor{gray}{(+4.33)} & 34.85 \textcolor{gray}{(+3.62)} & 16.71 \textcolor{gray}{(+1.77)} \\
    \midrule
    \textbf{DUSA (Ours)} & \textbf{43.61 \textcolor{gray}{(+8.88)}} & \textbf{38.46 \textcolor{gray}{(+7.23)}} & \textbf{20.09 \textcolor{gray}{(+5.15)}} \\
  \bottomrule
\end{tabular}
\end{table}

\subsubsection{Location-adaptive Sim2Real Adaptation}
To validate the effectiveness of the LSA, we temporarily remove the CIA and ablate the Location-adaptive Feature Selector (LFS). 
The results are shown in \Cref{tab:ablations}. 
\textit{+ LSA w/o LFS} means that the feature map is aggregated by a simple global average pooling layer instead of the LFS module. 
By adding the LFS module, the AP @ 0.7 improves from $15.76\%$ to $17.1\%$ ($+1.34\%$), which demonstrates that by adaptively selecting features of great importance according to the significance distribution, the sim2real domain gap could be better eliminated and the localization accuracy is significantly improved.

\subsubsection{Confidence-aware Inter-agent Adaptation}
We further validate the effectiveness of the CIA by temporarily removing the LSA and conducting ablations on the agent-wise confidence maps. 
The results are displayed in \Cref{tab:ablations}. 
\textit{+ CIA w/o Conf} means that the inter-agent discrimination loss is applied without the guidance of agent-wise confidence maps. 
By reweighing the inter-agent discrimination loss under the guidance of agent-wise confidence maps, the AP @ 0.7 improves from $16.14\%$ to $16.71\%$ ($+0.57\%$). 
This result proves the effectiveness of rebalancing the grids in the inter-agent discrimination loss.

Finally, by jointly applying the two adapters, \ie ~ LSA and CIA, the AP scores are further improved significantly. 
This strongly demonstrates the effectiveness of our decoupling strategy.

\subsection{Qualitative Evaluation}

We visualize the detection results of the V2X-ViT model without domain adaptation and the DUSA-adapted V2X-ViT model in \Cref{fig:vis}. 
In \Cref{fig:vis} (a) and \Cref{fig:vis} (b), we could observe that by applying DUSA to the collaborative detector, many of the missed cars can be detected. 
In \Cref{fig:vis} (c), we may also observe that by applying DUSA, the false positives are effectively suppressed. 
These observations indicate that the inter-domain feature maps are better aligned, and further prove the effectiveness of the proposed DUSA.

\section{Conclusion}
In this work, we propose a new unsupervised sim2real domain adaptation method for V2X collaborative detection named Decoupled Unsupervised Sim2Real Adaptation (DUSA). 
It decouples the V2X collaborative sim2real domain adaptation problem into two sub-problems: sim2real adaptation and inter-agent adaptation, and addresses them with the LSA module and the CIA module, respectively. 
The LSA module utilizes location significance to encourage the feature extractor to produce sim/real-invariant features. 
The CIA module leverages confidence clues to help make the feature extractor output agent-invariant features.
The two modules eliminate the sim/real and inter-agent domain gaps through adversarial training with GRLs.
Experiments on multiple collaborative 3D detectors demonstrate the effectiveness of the proposed DUSA method on unsupervised sim2real adaptation from the simulated V2XSet dataset to the real-world DAIR-V2X-C dataset.


\begin{acks}
This work was supported in part by the National Key R\&D Program of China under Grant 2022ZD0115502, in part by the National Natural Science Foundation of China under Grant 62122010, and in part by the Fundamental Research Funds for the Central Universities. 
\end{acks}

\bibliographystyle{ACM-Reference-Format}
\balance
\bibliography{base}


\begin{thebibliography}{54}


\ifx \showCODEN    \undefined \def \showCODEN     #1{\unskip}     \fi
\ifx \showDOI      \undefined \def \showDOI       #1{#1}\fi
\ifx \showISBNx    \undefined \def \showISBNx     #1{\unskip}     \fi
\ifx \showISBNxiii \undefined \def \showISBNxiii  #1{\unskip}     \fi
\ifx \showISSN     \undefined \def \showISSN      #1{\unskip}     \fi
\ifx \showLCCN     \undefined \def \showLCCN      #1{\unskip}     \fi
\ifx \shownote     \undefined \def \shownote      #1{#1}          \fi
\ifx \showarticletitle \undefined \def \showarticletitle #1{#1}   \fi
\ifx \showURL      \undefined \def \showURL       {\relax}        \fi
\providecommand\bibfield[2]{#2}
\providecommand\bibinfo[2]{#2}
\providecommand\natexlab[1]{#1}
\providecommand\showeprint[2][]{arXiv:#2}

\bibitem[Arnold et~al\mbox{.}(2020)]%
        {arnold2020cooperative}
\bibfield{author}{\bibinfo{person}{Eduardo Arnold}, \bibinfo{person}{Mehrdad
  Dianati}, \bibinfo{person}{Robert de Temple}, {and} \bibinfo{person}{Saber
  Fallah}.} \bibinfo{year}{2020}\natexlab{}.
\newblock \showarticletitle{Cooperative perception for 3D object detection in
  driving scenarios using infrastructure sensors}.
\newblock \bibinfo{journal}{\emph{IEEE Transactions on Intelligent
  Transportation Systems}} \bibinfo{volume}{23}, \bibinfo{number}{3}
  (\bibinfo{year}{2020}), \bibinfo{pages}{1852--1864}.
\newblock


\bibitem[Ben-David et~al\mbox{.}(2010)]%
        {ben2010theory}
\bibfield{author}{\bibinfo{person}{Shai Ben-David}, \bibinfo{person}{John
  Blitzer}, \bibinfo{person}{Koby Crammer}, \bibinfo{person}{Alex Kulesza},
  \bibinfo{person}{Fernando Pereira}, {and} \bibinfo{person}{Jennifer~Wortman
  Vaughan}.} \bibinfo{year}{2010}\natexlab{}.
\newblock \showarticletitle{A theory of learning from different domains}.
\newblock \bibinfo{journal}{\emph{Machine learning}}  \bibinfo{volume}{79}
  (\bibinfo{year}{2010}), \bibinfo{pages}{151--175}.
\newblock


\bibitem[Chen et~al\mbox{.}(2019a)]%
        {fcooper}
\bibfield{author}{\bibinfo{person}{Qi Chen}, \bibinfo{person}{Xu Ma},
  \bibinfo{person}{Sihai Tang}, \bibinfo{person}{Jingda Guo},
  \bibinfo{person}{Qing Yang}, {and} \bibinfo{person}{Song Fu}.}
  \bibinfo{year}{2019}\natexlab{a}.
\newblock \showarticletitle{F-Cooper: Feature Based Cooperative Perception for
  Autonomous Vehicle Edge Computing System Using 3D Point Clouds}. In
  \bibinfo{booktitle}{\emph{Proceedings of the 4th ACM/IEEE Symposium on Edge
  Computing}} (Arlington, Virginia) \emph{(\bibinfo{series}{SEC '19})}.
  \bibinfo{publisher}{Association for Computing Machinery},
  \bibinfo{address}{New York, NY, USA}, \bibinfo{pages}{88–100}.
\newblock
\showISBNx{9781450367332}
\urldef\tempurl%
\url{https://doi.org/10.1145/3318216.3363300}
\showDOI{\tempurl}


\bibitem[Chen et~al\mbox{.}(2019b)]%
        {cooper}
\bibfield{author}{\bibinfo{person}{Qi Chen}, \bibinfo{person}{Sihai Tang},
  \bibinfo{person}{Qing Yang}, {and} \bibinfo{person}{Song Fu}.}
  \bibinfo{year}{2019}\natexlab{b}.
\newblock \showarticletitle{Cooper: Cooperative Perception for Connected
  Autonomous Vehicles Based on 3D Point Clouds}. In
  \bibinfo{booktitle}{\emph{2019 IEEE 39th International Conference on
  Distributed Computing Systems (ICDCS)}}. \bibinfo{pages}{514--524}.
\newblock
\urldef\tempurl%
\url{https://doi.org/10.1109/ICDCS.2019.00058}
\showDOI{\tempurl}


\bibitem[Chen et~al\mbox{.}(2018)]%
        {chen2018domain}
\bibfield{author}{\bibinfo{person}{Yuhua Chen}, \bibinfo{person}{Wen Li},
  \bibinfo{person}{Christos Sakaridis}, \bibinfo{person}{Dengxin Dai}, {and}
  \bibinfo{person}{Luc Van~Gool}.} \bibinfo{year}{2018}\natexlab{}.
\newblock \showarticletitle{Domain adaptive faster r-cnn for object detection
  in the wild}. In \bibinfo{booktitle}{\emph{Proceedings of the IEEE conference
  on computer vision and pattern recognition}}. \bibinfo{pages}{3339--3348}.
\newblock


\bibitem[Deng et~al\mbox{.}(2021)]%
        {ic2fa}
\bibfield{author}{\bibinfo{person}{Wanxia Deng}, \bibinfo{person}{Yawen Cui},
  \bibinfo{person}{Zhen Liu}, \bibinfo{person}{Gangyao Kuang},
  \bibinfo{person}{Dewen Hu}, \bibinfo{person}{Matti Pietik\"{a}inen}, {and}
  \bibinfo{person}{Li Liu}.} \bibinfo{year}{2021}\natexlab{}.
\newblock \showarticletitle{Informative Class-Conditioned Feature Alignment for
  Unsupervised Domain Adaptation}. In \bibinfo{booktitle}{\emph{Proceedings of
  the 29th ACM International Conference on Multimedia}} (Virtual Event, China)
  \emph{(\bibinfo{series}{MM '21})}. \bibinfo{publisher}{Association for
  Computing Machinery}, \bibinfo{address}{New York, NY, USA},
  \bibinfo{pages}{1303–1312}.
\newblock
\showISBNx{9781450386517}
\urldef\tempurl%
\url{https://doi.org/10.1145/3474085.3475579}
\showDOI{\tempurl}


\bibitem[Dosovitskiy et~al\mbox{.}(2017)]%
        {carla}
\bibfield{author}{\bibinfo{person}{Alexey Dosovitskiy}, \bibinfo{person}{German
  Ros}, \bibinfo{person}{Felipe Codevilla}, \bibinfo{person}{Antonio Lopez},
  {and} \bibinfo{person}{Vladlen Koltun}.} \bibinfo{year}{2017}\natexlab{}.
\newblock \showarticletitle{{CARLA}: {An} Open Urban Driving Simulator}. In
  \bibinfo{booktitle}{\emph{Proceedings of the 1st Annual Conference on Robot
  Learning}} \emph{(\bibinfo{series}{Proceedings of Machine Learning Research},
  Vol.~\bibinfo{volume}{78})}, \bibfield{editor}{\bibinfo{person}{Sergey
  Levine}, \bibinfo{person}{Vincent Vanhoucke}, {and} \bibinfo{person}{Ken
  Goldberg}} (Eds.). \bibinfo{publisher}{PMLR}, \bibinfo{pages}{1--16}.
\newblock
\urldef\tempurl%
\url{https://proceedings.mlr.press/v78/dosovitskiy17a.html}
\showURL{%
\tempurl}


\bibitem[Ganin and Lempitsky(2015)]%
        {ganin2015unsupervised}
\bibfield{author}{\bibinfo{person}{Yaroslav Ganin} {and}
  \bibinfo{person}{Victor Lempitsky}.} \bibinfo{year}{2015}\natexlab{}.
\newblock \showarticletitle{Unsupervised domain adaptation by backpropagation}.
  In \bibinfo{booktitle}{\emph{International conference on machine learning}}.
  PMLR, \bibinfo{pages}{1180--1189}.
\newblock


\bibitem[Ganin et~al\mbox{.}(2016a)]%
        {ganin2016domain}
\bibfield{author}{\bibinfo{person}{Yaroslav Ganin}, \bibinfo{person}{Evgeniya
  Ustinova}, \bibinfo{person}{Hana Ajakan}, \bibinfo{person}{Pascal Germain},
  \bibinfo{person}{Hugo Larochelle}, \bibinfo{person}{Fran{\c{c}}ois
  Laviolette}, \bibinfo{person}{Mario Marchand}, {and} \bibinfo{person}{Victor
  Lempitsky}.} \bibinfo{year}{2016}\natexlab{a}.
\newblock \showarticletitle{Domain-adversarial training of neural networks}.
\newblock \bibinfo{journal}{\emph{The journal of machine learning research}}
  \bibinfo{volume}{17}, \bibinfo{number}{1} (\bibinfo{year}{2016}),
  \bibinfo{pages}{2096--2030}.
\newblock


\bibitem[Ganin et~al\mbox{.}(2016b)]%
        {grl}
\bibfield{author}{\bibinfo{person}{Yaroslav Ganin}, \bibinfo{person}{Evgeniya
  Ustinova}, \bibinfo{person}{Hana Ajakan}, \bibinfo{person}{Pascal Germain},
  \bibinfo{person}{Hugo Larochelle}, \bibinfo{person}{Fran{\c{c}}ois
  Laviolette}, \bibinfo{person}{Mario Marchand}, {and} \bibinfo{person}{Victor
  Lempitsky}.} \bibinfo{year}{2016}\natexlab{b}.
\newblock \showarticletitle{Domain-adversarial training of neural networks}.
\newblock \bibinfo{journal}{\emph{The journal of machine learning research}}
  \bibinfo{volume}{17}, \bibinfo{number}{1} (\bibinfo{year}{2016}),
  \bibinfo{pages}{2096--2030}.
\newblock


\bibitem[Goodfellow et~al\mbox{.}(2014)]%
        {NIPS2014_5ca3e9b1}
\bibfield{author}{\bibinfo{person}{Ian Goodfellow}, \bibinfo{person}{Jean
  Pouget-Abadie}, \bibinfo{person}{Mehdi Mirza}, \bibinfo{person}{Bing Xu},
  \bibinfo{person}{David Warde-Farley}, \bibinfo{person}{Sherjil Ozair},
  \bibinfo{person}{Aaron Courville}, {and} \bibinfo{person}{Yoshua Bengio}.}
  \bibinfo{year}{2014}\natexlab{}.
\newblock \showarticletitle{Generative Adversarial Nets}. In
  \bibinfo{booktitle}{\emph{Advances in Neural Information Processing
  Systems}}, \bibfield{editor}{\bibinfo{person}{Z.~Ghahramani},
  \bibinfo{person}{M.~Welling}, \bibinfo{person}{C.~Cortes},
  \bibinfo{person}{N.~Lawrence}, {and} \bibinfo{person}{K.Q. Weinberger}}
  (Eds.), Vol.~\bibinfo{volume}{27}. \bibinfo{publisher}{Curran Associates,
  Inc.}
\newblock
\urldef\tempurl%
\url{https://proceedings.neurips.cc/paper_files/paper/2014/file/5ca3e9b122f61f8f06494c97b1afccf3-Paper.pdf}
\showURL{%
\tempurl}


\bibitem[Gulrajani et~al\mbox{.}(2017)]%
        {gulrajani2017improved}
\bibfield{author}{\bibinfo{person}{Ishaan Gulrajani}, \bibinfo{person}{Faruk
  Ahmed}, \bibinfo{person}{Martin Arjovsky}, \bibinfo{person}{Vincent
  Dumoulin}, {and} \bibinfo{person}{Aaron~C Courville}.}
  \bibinfo{year}{2017}\natexlab{}.
\newblock \showarticletitle{Improved training of wasserstein gans}.
\newblock \bibinfo{journal}{\emph{Advances in neural information processing
  systems}}  \bibinfo{volume}{30} (\bibinfo{year}{2017}).
\newblock


\bibitem[Hobert et~al\mbox{.}(2015)]%
        {hobert2015enhancements}
\bibfield{author}{\bibinfo{person}{Laurens Hobert}, \bibinfo{person}{Andreas
  Festag}, \bibinfo{person}{Ignacio Llatser}, \bibinfo{person}{Luciano
  Altomare}, \bibinfo{person}{Filippo Visintainer}, {and}
  \bibinfo{person}{Andras Kovacs}.} \bibinfo{year}{2015}\natexlab{}.
\newblock \showarticletitle{Enhancements of V2X communication in support of
  cooperative autonomous driving}.
\newblock \bibinfo{journal}{\emph{IEEE communications magazine}}
  \bibinfo{volume}{53}, \bibinfo{number}{12} (\bibinfo{year}{2015}),
  \bibinfo{pages}{64--70}.
\newblock


\bibitem[Hoffman et~al\mbox{.}(2016)]%
        {hoffman2016fcns}
\bibfield{author}{\bibinfo{person}{Judy Hoffman}, \bibinfo{person}{Dequan
  Wang}, \bibinfo{person}{Fisher Yu}, {and} \bibinfo{person}{Trevor Darrell}.}
  \bibinfo{year}{2016}\natexlab{}.
\newblock \showarticletitle{Fcns in the wild: Pixel-level adversarial and
  constraint-based adaptation}.
\newblock \bibinfo{journal}{\emph{arXiv preprint arXiv:1612.02649}}
  (\bibinfo{year}{2016}).
\newblock


\bibitem[Hu et~al\mbox{.}(2022)]%
        {Where2comm:22}
\bibfield{author}{\bibinfo{person}{Yue Hu}, \bibinfo{person}{Shaoheng Fang},
  \bibinfo{person}{Zixing Lei}, \bibinfo{person}{Yiqi Zhong}, {and}
  \bibinfo{person}{Siheng Chen}.} \bibinfo{year}{2022}\natexlab{}.
\newblock \showarticletitle{Where2comm: Communication-Efficient Collaborative
  Perception via Spatial Confidence Maps}. In
  \bibinfo{booktitle}{\emph{Thirty-sixth Conference on Neural Information
  Processing Systems (Neurips)}}.
\newblock


\bibitem[Huang et~al\mbox{.}(2021)]%
        {imse}
\bibfield{author}{\bibinfo{person}{Shengqi Huang}, \bibinfo{person}{Wanqi
  Yang}, \bibinfo{person}{Lei Wang}, \bibinfo{person}{Luping Zhou}, {and}
  \bibinfo{person}{Ming Yang}.} \bibinfo{year}{2021}\natexlab{}.
\newblock \showarticletitle{Few-Shot Unsupervised Domain Adaptation with
  Image-to-Class Sparse Similarity Encoding}. In
  \bibinfo{booktitle}{\emph{Proceedings of the 29th ACM International
  Conference on Multimedia}} (Virtual Event, China) \emph{(\bibinfo{series}{MM
  '21})}. \bibinfo{publisher}{Association for Computing Machinery},
  \bibinfo{address}{New York, NY, USA}, \bibinfo{pages}{677–685}.
\newblock
\showISBNx{9781450386517}
\urldef\tempurl%
\url{https://doi.org/10.1145/3474085.3475232}
\showDOI{\tempurl}


\bibitem[Khodabandeh et~al\mbox{.}(2019)]%
        {khodabandeh2019robust}
\bibfield{author}{\bibinfo{person}{Mehran Khodabandeh}, \bibinfo{person}{Arash
  Vahdat}, \bibinfo{person}{Mani Ranjbar}, {and} \bibinfo{person}{William~G
  Macready}.} \bibinfo{year}{2019}\natexlab{}.
\newblock \showarticletitle{A robust learning approach to domain adaptive
  object detection}. In \bibinfo{booktitle}{\emph{Proceedings of the IEEE/CVF
  International Conference on Computer Vision}}. \bibinfo{pages}{480--490}.
\newblock


\bibitem[Krajzewicz et~al\mbox{.}(2012)]%
        {sumo}
\bibfield{author}{\bibinfo{person}{Daniel Krajzewicz}, \bibinfo{person}{Jakob
  Erdmann}, \bibinfo{person}{Michael Behrisch}, {and} \bibinfo{person}{Laura
  Bieker}.} \bibinfo{year}{2012}\natexlab{}.
\newblock \showarticletitle{Recent development and applications of
  SUMO-Simulation of Urban MObility}.
\newblock \bibinfo{journal}{\emph{International journal on advances in systems
  and measurements}} \bibinfo{volume}{5}, \bibinfo{number}{3\&4}
  (\bibinfo{year}{2012}).
\newblock


\bibitem[Lang et~al\mbox{.}(2019)]%
        {pointpillars}
\bibfield{author}{\bibinfo{person}{Alex~H. Lang}, \bibinfo{person}{Sourabh
  Vora}, \bibinfo{person}{Holger Caesar}, \bibinfo{person}{Lubing Zhou},
  \bibinfo{person}{Jiong Yang}, {and} \bibinfo{person}{Oscar Beijbom}.}
  \bibinfo{year}{2019}\natexlab{}.
\newblock \showarticletitle{PointPillars: Fast Encoders for Object Detection
  From Point Clouds}. In \bibinfo{booktitle}{\emph{2019 IEEE/CVF Conference on
  Computer Vision and Pattern Recognition (CVPR)}}.
  \bibinfo{pages}{12689--12697}.
\newblock
\urldef\tempurl%
\url{https://doi.org/10.1109/CVPR.2019.01298}
\showDOI{\tempurl}


\bibitem[Lee et~al\mbox{.}(2013)]%
        {lee2013pseudo}
\bibfield{author}{\bibinfo{person}{Dong-Hyun Lee} {et~al\mbox{.}}}
  \bibinfo{year}{2013}\natexlab{}.
\newblock \showarticletitle{Pseudo-label: The simple and efficient
  semi-supervised learning method for deep neural networks}. In
  \bibinfo{booktitle}{\emph{Workshop on challenges in representation learning,
  ICML}}, Vol.~\bibinfo{volume}{3}. \bibinfo{pages}{896}.
\newblock


\bibitem[Lei et~al\mbox{.}(2022)]%
        {lei2022latency}
\bibfield{author}{\bibinfo{person}{Zixing Lei}, \bibinfo{person}{Shunli Ren},
  \bibinfo{person}{Yue Hu}, \bibinfo{person}{Wenjun Zhang}, {and}
  \bibinfo{person}{Siheng Chen}.} \bibinfo{year}{2022}\natexlab{}.
\newblock \showarticletitle{Latency-Aware Collaborative Perception}. In
  \bibinfo{booktitle}{\emph{Computer Vision -- ECCV 2022}},
  \bibfield{editor}{\bibinfo{person}{Shai Avidan}, \bibinfo{person}{Gabriel
  Brostow}, \bibinfo{person}{Moustapha Ciss{\'e}},
  \bibinfo{person}{Giovanni~Maria Farinella}, {and} \bibinfo{person}{Tal
  Hassner}} (Eds.). \bibinfo{publisher}{Springer Nature Switzerland},
  \bibinfo{address}{Cham}, \bibinfo{pages}{316--332}.
\newblock
\showISBNx{978-3-031-19824-3}


\bibitem[Li et~al\mbox{.}(2022)]%
        {v2xsim}
\bibfield{author}{\bibinfo{person}{Yiming Li}, \bibinfo{person}{Dekun Ma},
  \bibinfo{person}{Ziyan An}, \bibinfo{person}{Zixun Wang},
  \bibinfo{person}{Yiqi Zhong}, \bibinfo{person}{Siheng Chen}, {and}
  \bibinfo{person}{Chen Feng}.} \bibinfo{year}{2022}\natexlab{}.
\newblock \showarticletitle{V2X-Sim: Multi-Agent Collaborative Perception
  Dataset and Benchmark for Autonomous Driving}.
\newblock \bibinfo{journal}{\emph{IEEE Robotics and Automation Letters}}
  \bibinfo{volume}{7}, \bibinfo{number}{4} (\bibinfo{year}{2022}),
  \bibinfo{pages}{10914--10921}.
\newblock
\urldef\tempurl%
\url{https://doi.org/10.1109/LRA.2022.3192802}
\showDOI{\tempurl}


\bibitem[Li et~al\mbox{.}(2021)]%
        {disconet}
\bibfield{author}{\bibinfo{person}{Yiming Li}, \bibinfo{person}{Shunli Ren},
  \bibinfo{person}{Pengxiang Wu}, \bibinfo{person}{Siheng Chen},
  \bibinfo{person}{Chen Feng}, {and} \bibinfo{person}{Wenjun Zhang}.}
  \bibinfo{year}{2021}\natexlab{}.
\newblock \showarticletitle{Learning Distilled Collaboration Graph for
  Multi-Agent Perception}. In \bibinfo{booktitle}{\emph{Advances in Neural
  Information Processing Systems}},
  \bibfield{editor}{\bibinfo{person}{M.~Ranzato},
  \bibinfo{person}{A.~Beygelzimer}, \bibinfo{person}{Y.~Dauphin},
  \bibinfo{person}{P.S. Liang}, {and} \bibinfo{person}{J.~Wortman Vaughan}}
  (Eds.), Vol.~\bibinfo{volume}{34}. \bibinfo{publisher}{Curran Associates,
  Inc.}, \bibinfo{pages}{29541--29552}.
\newblock
\urldef\tempurl%
\url{https://proceedings.neurips.cc/paper_files/paper/2021/file/f702defbc67edb455949f46babab0c18-Paper.pdf}
\showURL{%
\tempurl}


\bibitem[Lin et~al\mbox{.}(2020)]%
        {focalloss}
\bibfield{author}{\bibinfo{person}{Tsung-Yi Lin}, \bibinfo{person}{Priya
  Goyal}, \bibinfo{person}{Ross Girshick}, \bibinfo{person}{Kaiming He}, {and}
  \bibinfo{person}{Piotr Dollár}.} \bibinfo{year}{2020}\natexlab{}.
\newblock \showarticletitle{Focal Loss for Dense Object Detection}.
\newblock \bibinfo{journal}{\emph{IEEE Transactions on Pattern Analysis and
  Machine Intelligence}} \bibinfo{volume}{42}, \bibinfo{number}{2}
  (\bibinfo{year}{2020}), \bibinfo{pages}{318--327}.
\newblock
\urldef\tempurl%
\url{https://doi.org/10.1109/TPAMI.2018.2858826}
\showDOI{\tempurl}


\bibitem[Liu et~al\mbox{.}(2021)]%
        {liu2021towards}
\bibfield{author}{\bibinfo{person}{Shaoshan Liu}, \bibinfo{person}{Bo Yu},
  \bibinfo{person}{Jie Tang}, {and} \bibinfo{person}{Qi Zhu}.}
  \bibinfo{year}{2021}\natexlab{}.
\newblock \showarticletitle{Towards fully intelligent transportation through
  infrastructure-vehicle cooperative autonomous driving: Challenges and
  opportunities}. In \bibinfo{booktitle}{\emph{2021 58th ACM/IEEE Design
  Automation Conference (DAC)}}. IEEE, \bibinfo{pages}{1323--1326}.
\newblock


\bibitem[Liu et~al\mbox{.}(2020a)]%
        {when2com}
\bibfield{author}{\bibinfo{person}{Yen-Cheng Liu}, \bibinfo{person}{Junjiao
  Tian}, \bibinfo{person}{Nathaniel Glaser}, {and} \bibinfo{person}{Zsolt
  Kira}.} \bibinfo{year}{2020}\natexlab{a}.
\newblock \showarticletitle{When2com: Multi-Agent Perception via Communication
  Graph Grouping}. In \bibinfo{booktitle}{\emph{2020 IEEE/CVF Conference on
  Computer Vision and Pattern Recognition (CVPR)}}.
  \bibinfo{pages}{4105--4114}.
\newblock
\urldef\tempurl%
\url{https://doi.org/10.1109/CVPR42600.2020.00416}
\showDOI{\tempurl}


\bibitem[Liu et~al\mbox{.}(2020b)]%
        {who2com}
\bibfield{author}{\bibinfo{person}{Yen-Cheng Liu}, \bibinfo{person}{Junjiao
  Tian}, \bibinfo{person}{Chih-Yao Ma}, \bibinfo{person}{Nathan Glaser},
  \bibinfo{person}{Chia-Wen Kuo}, {and} \bibinfo{person}{Zsolt Kira}.}
  \bibinfo{year}{2020}\natexlab{b}.
\newblock \showarticletitle{Who2com: Collaborative Perception via Learnable
  Handshake Communication}. In \bibinfo{booktitle}{\emph{2020 IEEE
  International Conference on Robotics and Automation (ICRA)}}.
  \bibinfo{pages}{6876--6883}.
\newblock
\urldef\tempurl%
\url{https://doi.org/10.1109/ICRA40945.2020.9197364}
\showDOI{\tempurl}


\bibitem[Lu et~al\mbox{.}(2023)]%
        {coalign}
\bibfield{author}{\bibinfo{person}{Yifan Lu}, \bibinfo{person}{Quanhao Li},
  \bibinfo{person}{Baoan Liu}, \bibinfo{person}{Mehrdad Dianati},
  \bibinfo{person}{Chen Feng}, \bibinfo{person}{Siheng Chen}, {and}
  \bibinfo{person}{Yanfeng Wang}.} \bibinfo{year}{2023}\natexlab{}.
\newblock \bibinfo{title}{Robust Collaborative 3D Object Detection in Presence
  of Pose Errors}.
\newblock
\newblock
\showeprint[arxiv]{2211.07214}~[cs.CV]


\bibitem[Luo et~al\mbox{.}(2022)]%
        {crcnet}
\bibfield{author}{\bibinfo{person}{Guiyang Luo}, \bibinfo{person}{Hui Zhang},
  \bibinfo{person}{Quan Yuan}, {and} \bibinfo{person}{Jinglin Li}.}
  \bibinfo{year}{2022}\natexlab{}.
\newblock \showarticletitle{Complementarity-Enhanced and Redundancy-Minimized
  Collaboration Network for Multi-Agent Perception}. In
  \bibinfo{booktitle}{\emph{Proceedings of the 30th ACM International
  Conference on Multimedia}} (Lisboa, Portugal) \emph{(\bibinfo{series}{MM
  '22})}. \bibinfo{publisher}{Association for Computing Machinery},
  \bibinfo{address}{New York, NY, USA}, \bibinfo{pages}{3578–3586}.
\newblock
\showISBNx{9781450392037}
\urldef\tempurl%
\url{https://doi.org/10.1145/3503161.3548197}
\showDOI{\tempurl}


\bibitem[Luo et~al\mbox{.}(2021)]%
        {luo2021unsupervised}
\bibfield{author}{\bibinfo{person}{Zhipeng Luo}, \bibinfo{person}{Zhongang
  Cai}, \bibinfo{person}{Changqing Zhou}, \bibinfo{person}{Gongjie Zhang},
  \bibinfo{person}{Haiyu Zhao}, \bibinfo{person}{Shuai Yi},
  \bibinfo{person}{Shijian Lu}, \bibinfo{person}{Hongsheng Li},
  \bibinfo{person}{Shanghang Zhang}, {and} \bibinfo{person}{Ziwei Liu}.}
  \bibinfo{year}{2021}\natexlab{}.
\newblock \showarticletitle{Unsupervised domain adaptive 3d detection with
  multi-level consistency}. In \bibinfo{booktitle}{\emph{Proceedings of the
  IEEE/CVF International Conference on Computer Vision}}.
  \bibinfo{pages}{8866--8875}.
\newblock


\bibitem[Matthew~Howe(2021)]%
        {WIBAM}
\bibfield{author}{\bibinfo{person}{Jamie~Mackenzie Matthew~Howe, Ian~Reid}.}
  \bibinfo{year}{2021}\natexlab{}.
\newblock \showarticletitle{Weakly Supervised Training of Monocular 3D Object
  Detectors Using Wide Baseline Multi-view Traffic Camera Data}.
\newblock \bibinfo{journal}{\emph{32nd British Machine Vision Conference, BMVC
  2021}} (\bibinfo{year}{2021}).
\newblock


\bibitem[Qiao and Zulkernine(2023)]%
        {Qiao_2023_WACV}
\bibfield{author}{\bibinfo{person}{Donghao Qiao} {and} \bibinfo{person}{Farhana
  Zulkernine}.} \bibinfo{year}{2023}\natexlab{}.
\newblock \showarticletitle{Adaptive Feature Fusion for Cooperative Perception
  Using LiDAR Point Clouds}. In \bibinfo{booktitle}{\emph{Proceedings of the
  IEEE/CVF Winter Conference on Applications of Computer Vision (WACV)}}.
  \bibinfo{pages}{1186--1195}.
\newblock


\bibitem[Qin et~al\mbox{.}(2019)]%
        {qin2019pointdan}
\bibfield{author}{\bibinfo{person}{Can Qin}, \bibinfo{person}{Haoxuan You},
  \bibinfo{person}{Lichen Wang}, \bibinfo{person}{C-C~Jay Kuo}, {and}
  \bibinfo{person}{Yun Fu}.} \bibinfo{year}{2019}\natexlab{}.
\newblock \showarticletitle{Pointdan: A multi-scale 3d domain adaption network
  for point cloud representation}.
\newblock \bibinfo{journal}{\emph{Advances in Neural Information Processing
  Systems}}  \bibinfo{volume}{32} (\bibinfo{year}{2019}).
\newblock


\bibitem[Saito et~al\mbox{.}(2017)]%
        {saito2017asymmetric}
\bibfield{author}{\bibinfo{person}{Kuniaki Saito}, \bibinfo{person}{Yoshitaka
  Ushiku}, {and} \bibinfo{person}{Tatsuya Harada}.}
  \bibinfo{year}{2017}\natexlab{}.
\newblock \showarticletitle{Asymmetric tri-training for unsupervised domain
  adaptation}. In \bibinfo{booktitle}{\emph{International Conference on Machine
  Learning}}. PMLR, \bibinfo{pages}{2988--2997}.
\newblock


\bibitem[Tsai et~al\mbox{.}(2018)]%
        {tsai2018learning}
\bibfield{author}{\bibinfo{person}{Yi-Hsuan Tsai}, \bibinfo{person}{Wei-Chih
  Hung}, \bibinfo{person}{Samuel Schulter}, \bibinfo{person}{Kihyuk Sohn},
  \bibinfo{person}{Ming-Hsuan Yang}, {and} \bibinfo{person}{Manmohan
  Chandraker}.} \bibinfo{year}{2018}\natexlab{}.
\newblock \showarticletitle{Learning to adapt structured output space for
  semantic segmentation}. In \bibinfo{booktitle}{\emph{Proceedings of the IEEE
  conference on computer vision and pattern recognition}}.
  \bibinfo{pages}{7472--7481}.
\newblock


\bibitem[Wang et~al\mbox{.}(2021)]%
        {interbn}
\bibfield{author}{\bibinfo{person}{Mengzhu Wang}, \bibinfo{person}{Wei Wang},
  \bibinfo{person}{Baopu Li}, \bibinfo{person}{Xiang Zhang},
  \bibinfo{person}{Long Lan}, \bibinfo{person}{Huibin Tan},
  \bibinfo{person}{Tianyi Liang}, \bibinfo{person}{Wei Yu}, {and}
  \bibinfo{person}{Zhigang Luo}.} \bibinfo{year}{2021}\natexlab{}.
\newblock \showarticletitle{InterBN: Channel Fusion for Adversarial
  Unsupervised Domain Adaptation}. In \bibinfo{booktitle}{\emph{Proceedings of
  the 29th ACM International Conference on Multimedia}} (Virtual Event, China)
  \emph{(\bibinfo{series}{MM '21})}. \bibinfo{publisher}{Association for
  Computing Machinery}, \bibinfo{address}{New York, NY, USA},
  \bibinfo{pages}{3691–3700}.
\newblock
\showISBNx{9781450386517}
\urldef\tempurl%
\url{https://doi.org/10.1145/3474085.3475481}
\showDOI{\tempurl}


\bibitem[Wang et~al\mbox{.}(2020b)]%
        {v2vnet}
\bibfield{author}{\bibinfo{person}{Tsun-Hsuan Wang}, \bibinfo{person}{Sivabalan
  Manivasagam}, \bibinfo{person}{Ming Liang}, \bibinfo{person}{Bin Yang},
  \bibinfo{person}{Wenyuan Zeng}, {and} \bibinfo{person}{Raquel Urtasun}.}
  \bibinfo{year}{2020}\natexlab{b}.
\newblock \showarticletitle{V2VNet: Vehicle-to-Vehicle Communication for Joint
  Perception and Prediction}. In \bibinfo{booktitle}{\emph{Computer Vision --
  ECCV 2020}}, \bibfield{editor}{\bibinfo{person}{Andrea Vedaldi},
  \bibinfo{person}{Horst Bischof}, \bibinfo{person}{Thomas Brox}, {and}
  \bibinfo{person}{Jan-Michael Frahm}} (Eds.). \bibinfo{publisher}{Springer
  International Publishing}, \bibinfo{address}{Cham},
  \bibinfo{pages}{605--621}.
\newblock
\showISBNx{978-3-030-58536-5}


\bibitem[Wang et~al\mbox{.}(2020a)]%
        {wang2020train}
\bibfield{author}{\bibinfo{person}{Yan Wang}, \bibinfo{person}{Xiangyu Chen},
  \bibinfo{person}{Yurong You}, \bibinfo{person}{Li~Erran Li},
  \bibinfo{person}{Bharath Hariharan}, \bibinfo{person}{Mark Campbell},
  \bibinfo{person}{Kilian~Q Weinberger}, {and} \bibinfo{person}{Wei-Lun Chao}.}
  \bibinfo{year}{2020}\natexlab{a}.
\newblock \showarticletitle{Train in germany, test in the usa: Making 3d object
  detectors generalize}. In \bibinfo{booktitle}{\emph{Proceedings of the
  IEEE/CVF Conference on Computer Vision and Pattern Recognition}}.
  \bibinfo{pages}{11713--11723}.
\newblock


\bibitem[Xu et~al\mbox{.}(2023a)]%
        {modelagnostic}
\bibfield{author}{\bibinfo{person}{Runsheng Xu}, \bibinfo{person}{Weizhe Chen},
  \bibinfo{person}{Hao Xiang}, \bibinfo{person}{Lantao Liu}, {and}
  \bibinfo{person}{Jiaqi Ma}.} \bibinfo{year}{2023}\natexlab{a}.
\newblock \bibinfo{title}{Model-Agnostic Multi-Agent Perception Framework}.
\newblock
\newblock
\showeprint[arxiv]{2203.13168}~[cs.RO]


\bibitem[Xu et~al\mbox{.}(2021)]%
        {opencda}
\bibfield{author}{\bibinfo{person}{Runsheng Xu}, \bibinfo{person}{Yi Guo},
  \bibinfo{person}{Xu Han}, \bibinfo{person}{Xin Xia}, \bibinfo{person}{Hao
  Xiang}, {and} \bibinfo{person}{Jiaqi Ma}.} \bibinfo{year}{2021}\natexlab{}.
\newblock \showarticletitle{OpenCDA: An Open Cooperative Driving Automation
  Framework Integrated with Co-Simulation}. In \bibinfo{booktitle}{\emph{2021
  IEEE International Intelligent Transportation Systems Conference (ITSC)}}.
  \bibinfo{pages}{1155--1162}.
\newblock
\urldef\tempurl%
\url{https://doi.org/10.1109/ITSC48978.2021.9564825}
\showDOI{\tempurl}


\bibitem[Xu et~al\mbox{.}(2023b)]%
        {xu2023mpda}
\bibfield{author}{\bibinfo{person}{Runsheng Xu}, \bibinfo{person}{Jinlong Li},
  \bibinfo{person}{Xiaoyu Dong}, \bibinfo{person}{Hongkai Yu}, {and}
  \bibinfo{person}{Jiaqi Ma}.} \bibinfo{year}{2023}\natexlab{b}.
\newblock \showarticletitle{Bridging the Domain Gap for Multi-Agent
  Perception}. In \bibinfo{booktitle}{\emph{2023 IEEE International Conference
  on Robotics and Automation (ICRA)}}.
\newblock


\bibitem[Xu et~al\mbox{.}(2022a)]%
        {xu2022cobevt}
\bibfield{author}{\bibinfo{person}{Runsheng Xu}, \bibinfo{person}{Zhengzhong
  Tu}, \bibinfo{person}{Hao Xiang}, \bibinfo{person}{Wei Shao},
  \bibinfo{person}{Bolei Zhou}, {and} \bibinfo{person}{Jiaqi Ma}.}
  \bibinfo{year}{2022}\natexlab{a}.
\newblock \showarticletitle{CoBEVT: Cooperative Bird's Eye View Semantic
  Segmentation with Sparse Transformers}. In
  \bibinfo{booktitle}{\emph{Conference on Robot Learning (CoRL)}}.
\newblock


\bibitem[Xu et~al\mbox{.}(2022b)]%
        {v2xvit}
\bibfield{author}{\bibinfo{person}{Runsheng Xu}, \bibinfo{person}{Hao Xiang},
  \bibinfo{person}{Zhengzhong Tu}, \bibinfo{person}{Xin Xia},
  \bibinfo{person}{Ming-Hsuan Yang}, {and} \bibinfo{person}{Jiaqi Ma}.}
  \bibinfo{year}{2022}\natexlab{b}.
\newblock \showarticletitle{V2X-ViT: Vehicle-to-Everything Cooperative
  Perception with Vision Transformer}. In \bibinfo{booktitle}{\emph{Computer
  Vision -- ECCV 2022}}, \bibfield{editor}{\bibinfo{person}{Shai Avidan},
  \bibinfo{person}{Gabriel Brostow}, \bibinfo{person}{Moustapha Ciss{\'e}},
  \bibinfo{person}{Giovanni~Maria Farinella}, {and} \bibinfo{person}{Tal
  Hassner}} (Eds.). \bibinfo{publisher}{Springer Nature Switzerland},
  \bibinfo{address}{Cham}, \bibinfo{pages}{107--124}.
\newblock
\showISBNx{978-3-031-19842-7}


\bibitem[Xu et~al\mbox{.}(2022c)]%
        {opv2v}
\bibfield{author}{\bibinfo{person}{Runsheng Xu}, \bibinfo{person}{Hao Xiang},
  \bibinfo{person}{Xin Xia}, \bibinfo{person}{Xu Han}, \bibinfo{person}{Jinlong
  Li}, {and} \bibinfo{person}{Jiaqi Ma}.} \bibinfo{year}{2022}\natexlab{c}.
\newblock \showarticletitle{OPV2V: An Open Benchmark Dataset and Fusion
  Pipeline for Perception with Vehicle-to-Vehicle Communication}. In
  \bibinfo{booktitle}{\emph{2022 International Conference on Robotics and
  Automation (ICRA)}}. \bibinfo{pages}{2583--2589}.
\newblock
\urldef\tempurl%
\url{https://doi.org/10.1109/ICRA46639.2022.9812038}
\showDOI{\tempurl}


\bibitem[Yang et~al\mbox{.}(2021)]%
        {Yang_2021_CVPR}
\bibfield{author}{\bibinfo{person}{Jihan Yang}, \bibinfo{person}{Shaoshuai
  Shi}, \bibinfo{person}{Zhe Wang}, \bibinfo{person}{Hongsheng Li}, {and}
  \bibinfo{person}{Xiaojuan Qi}.} \bibinfo{year}{2021}\natexlab{}.
\newblock \showarticletitle{ST3D: Self-Training for Unsupervised Domain
  Adaptation on 3D Object Detection}. In \bibinfo{booktitle}{\emph{Proceedings
  of the IEEE/CVF Conference on Computer Vision and Pattern Recognition
  (CVPR)}}. \bibinfo{pages}{10368--10378}.
\newblock


\bibitem[Yang et~al\mbox{.}(2022)]%
        {yang2022st3d++}
\bibfield{author}{\bibinfo{person}{Jihan Yang}, \bibinfo{person}{Shaoshuai
  Shi}, \bibinfo{person}{Zhe Wang}, \bibinfo{person}{Hongsheng Li}, {and}
  \bibinfo{person}{Xiaojuan Qi}.} \bibinfo{year}{2022}\natexlab{}.
\newblock \showarticletitle{ST3D++: denoised self-training for unsupervised
  domain adaptation on 3D object detection}.
\newblock \bibinfo{journal}{\emph{IEEE Transactions on Pattern Analysis and
  Machine Intelligence}} (\bibinfo{year}{2022}).
\newblock


\bibitem[Ye et~al\mbox{.}(2021)]%
        {sdfuda}
\bibfield{author}{\bibinfo{person}{Mucong Ye}, \bibinfo{person}{Jing Zhang},
  \bibinfo{person}{Jinpeng Ouyang}, {and} \bibinfo{person}{Ding Yuan}.}
  \bibinfo{year}{2021}\natexlab{}.
\newblock \showarticletitle{Source Data-Free Unsupervised Domain Adaptation for
  Semantic Segmentation}. In \bibinfo{booktitle}{\emph{Proceedings of the 29th
  ACM International Conference on Multimedia}} (Virtual Event, China)
  \emph{(\bibinfo{series}{MM '21})}. \bibinfo{publisher}{Association for
  Computing Machinery}, \bibinfo{address}{New York, NY, USA},
  \bibinfo{pages}{2233–2242}.
\newblock
\showISBNx{9781450386517}
\urldef\tempurl%
\url{https://doi.org/10.1145/3474085.3475384}
\showDOI{\tempurl}


\bibitem[Yu et~al\mbox{.}(2022)]%
        {dair}
\bibfield{author}{\bibinfo{person}{Haibao Yu}, \bibinfo{person}{Yizhen Luo},
  \bibinfo{person}{Mao Shu}, \bibinfo{person}{Yiyi Huo},
  \bibinfo{person}{Zebang Yang}, \bibinfo{person}{Yifeng Shi},
  \bibinfo{person}{Zhenglong Guo}, \bibinfo{person}{Hanyu Li},
  \bibinfo{person}{Xing Hu}, \bibinfo{person}{Jirui Yuan}, {and}
  \bibinfo{person}{Zaiqing Nie}.} \bibinfo{year}{2022}\natexlab{}.
\newblock \showarticletitle{DAIR-V2X: A Large-Scale Dataset for
  Vehicle-Infrastructure Cooperative 3D Object Detection}. In
  \bibinfo{booktitle}{\emph{2022 IEEE/CVF Conference on Computer Vision and
  Pattern Recognition (CVPR)}}. \bibinfo{pages}{21329--21338}.
\newblock
\urldef\tempurl%
\url{https://doi.org/10.1109/CVPR52688.2022.02067}
\showDOI{\tempurl}


\bibitem[Yuan et~al\mbox{.}(2022)]%
        {fpvrcnn}
\bibfield{author}{\bibinfo{person}{Yunshuang Yuan}, \bibinfo{person}{Hao
  Cheng}, {and} \bibinfo{person}{Monika Sester}.}
  \bibinfo{year}{2022}\natexlab{}.
\newblock \showarticletitle{Keypoints-Based Deep Feature Fusion for Cooperative
  Vehicle Detection of Autonomous Driving}.
\newblock \bibinfo{journal}{\emph{IEEE Robotics and Automation Letters}}
  \bibinfo{volume}{7}, \bibinfo{number}{2} (\bibinfo{year}{2022}),
  \bibinfo{pages}{3054--3061}.
\newblock
\urldef\tempurl%
\url{https://doi.org/10.1109/LRA.2022.3143299}
\showDOI{\tempurl}


\bibitem[Zhang et~al\mbox{.}(2021b)]%
        {m2guda}
\bibfield{author}{\bibinfo{person}{Chengyuan Zhang}, \bibinfo{person}{Zhi
  Zhong}, \bibinfo{person}{Lei Zhu}, \bibinfo{person}{Shichao Zhang},
  \bibinfo{person}{Da Cao}, {and} \bibinfo{person}{Jianfeng Zhang}.}
  \bibinfo{year}{2021}\natexlab{b}.
\newblock \showarticletitle{M2GUDA: Multi-Metrics Graph-Based Unsupervised
  Domain Adaptation for Cross-Modal Hashing}. In
  \bibinfo{booktitle}{\emph{Proceedings of the 2021 International Conference on
  Multimedia Retrieval}} (Taipei, Taiwan) \emph{(\bibinfo{series}{ICMR '21})}.
  \bibinfo{publisher}{Association for Computing Machinery},
  \bibinfo{address}{New York, NY, USA}, \bibinfo{pages}{674–681}.
\newblock
\showISBNx{9781450384636}
\urldef\tempurl%
\url{https://doi.org/10.1145/3460426.3463670}
\showDOI{\tempurl}


\bibitem[Zhang et~al\mbox{.}(2021a)]%
        {zhang2021srdan}
\bibfield{author}{\bibinfo{person}{Weichen Zhang}, \bibinfo{person}{Wen Li},
  {and} \bibinfo{person}{Dong Xu}.} \bibinfo{year}{2021}\natexlab{a}.
\newblock \showarticletitle{SRDAN: Scale-aware and range-aware domain
  adaptation network for cross-dataset 3D object detection}. In
  \bibinfo{booktitle}{\emph{Proceedings of the IEEE/CVF Conference on Computer
  Vision and Pattern Recognition}}. \bibinfo{pages}{6769--6779}.
\newblock


\bibitem[Zhao et~al\mbox{.}(2018)]%
        {emotiongan}
\bibfield{author}{\bibinfo{person}{Sicheng Zhao}, \bibinfo{person}{Xin Zhao},
  \bibinfo{person}{Guiguang Ding}, {and} \bibinfo{person}{Kurt Keutzer}.}
  \bibinfo{year}{2018}\natexlab{}.
\newblock \showarticletitle{EmotionGAN: Unsupervised Domain Adaptation for
  Learning Discrete Probability Distributions of Image Emotions}. In
  \bibinfo{booktitle}{\emph{Proceedings of the 26th ACM International
  Conference on Multimedia}} (Seoul, Republic of Korea)
  \emph{(\bibinfo{series}{MM '18})}. \bibinfo{publisher}{Association for
  Computing Machinery}, \bibinfo{address}{New York, NY, USA},
  \bibinfo{pages}{1319–1327}.
\newblock
\showISBNx{9781450356657}
\urldef\tempurl%
\url{https://doi.org/10.1145/3240508.3240591}
\showDOI{\tempurl}


\bibitem[Zhou et~al\mbox{.}(2022)]%
        {cdal}
\bibfield{author}{\bibinfo{person}{Lihua Zhou}, \bibinfo{person}{Mao Ye},
  \bibinfo{person}{Xiatian Zhu}, \bibinfo{person}{Shuaifeng Li}, {and}
  \bibinfo{person}{Yiguang Liu}.} \bibinfo{year}{2022}\natexlab{}.
\newblock \showarticletitle{Class Discriminative Adversarial Learning for
  Unsupervised Domain Adaptation}. In \bibinfo{booktitle}{\emph{Proceedings of
  the 30th ACM International Conference on Multimedia}} (Lisboa, Portugal)
  \emph{(\bibinfo{series}{MM '22})}. \bibinfo{publisher}{Association for
  Computing Machinery}, \bibinfo{address}{New York, NY, USA},
  \bibinfo{pages}{4318–4326}.
\newblock
\showISBNx{9781450392037}
\urldef\tempurl%
\url{https://doi.org/10.1145/3503161.3548143}
\showDOI{\tempurl}


\bibitem[Zou et~al\mbox{.}(2018)]%
        {zou2018unsupervised}
\bibfield{author}{\bibinfo{person}{Yang Zou}, \bibinfo{person}{Zhiding Yu},
  \bibinfo{person}{BVK Kumar}, {and} \bibinfo{person}{Jinsong Wang}.}
  \bibinfo{year}{2018}\natexlab{}.
\newblock \showarticletitle{Unsupervised domain adaptation for semantic
  segmentation via class-balanced self-training}. In
  \bibinfo{booktitle}{\emph{Proceedings of the European conference on computer
  vision (ECCV)}}. \bibinfo{pages}{289--305}.
\newblock


\end{thebibliography}

\newpage
\appendix

\section{More Analysis on Inter-agent Domain Gap}

\begin{figure}[ht]
  \centering
  \includegraphics[width=0.85\linewidth]{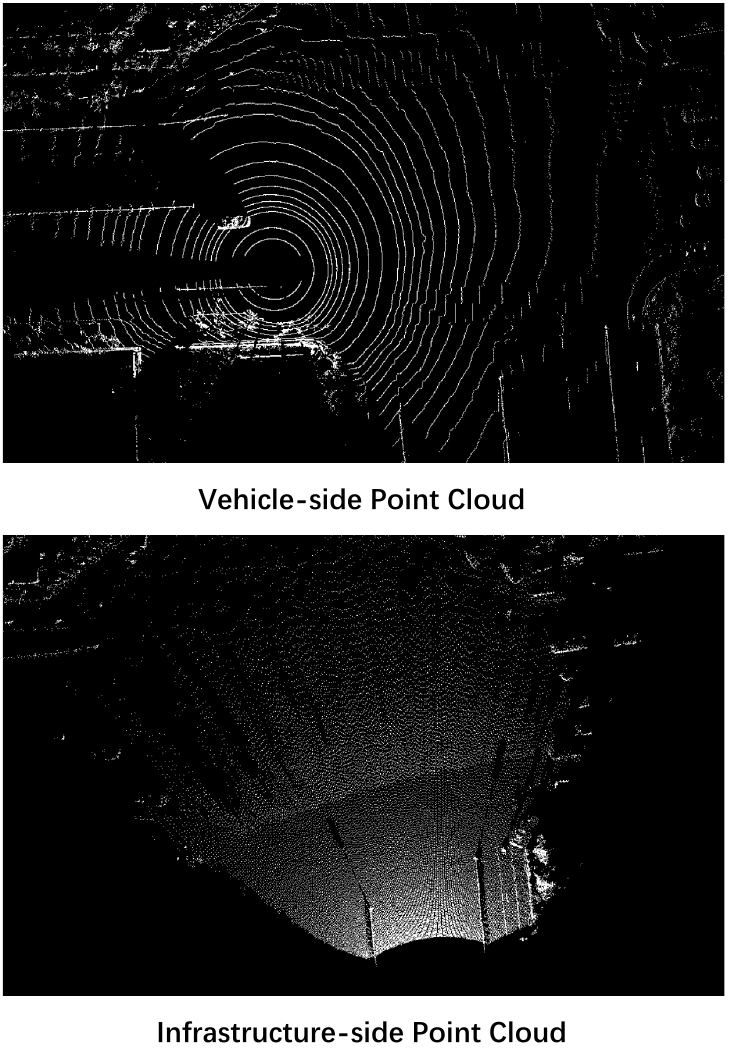}
  \caption{Comparison of a vehicle-side point cloud and an infrastructure-side point cloud from DAIR-V2X-C \cite{dair}.}
  \Description{Comparison of point clouds from heterogeneous agents.}
  \label{fig:agent_comp}
\end{figure}

In real-world V2X applications, the collaborative agents are usually heterogeneous in sensor type and pose, \etc ~
For instance, DAIR-V2X-C \cite{dair} contains two types of collaborative agents: data acquisition cars and roadside infrastructures. 
Different types of LiDARs are installed on cars and infrastructures, as shown in \Cref{tab:sensors}. 
In addition, the installation height of infrastructure-side LiDARs is usually higher than that of vehicle-side LiDARs. 
Moreover, infrastructure-side LiDARs often have a larger pitch angle than vehicle-side LiDARs. 
The differences in LiDAR type and placement result in divergence in point clouds' density, distribution, range, reflectance, \etc, inducing the inter-agent domain gap. 
We visualize a vehicle-side point cloud and an infrastructure-side point cloud from the validation split of the DAIR-V2X-C \cite{dair} dataset for comparison in \Cref{fig:agent_comp} to demonstrate the inter-agent domain gap.

\begin{table}
  \caption{Comparison of LiDARs on different types of collaborative agents in DAIR-V2X-C \cite{dair}.}
  \label{tab:sensors}
  \begin{tabular}{ccccc}
    \toprule
    Agent & Type & Beams & \makecell[c]{Horizontal \\ FoV} & Range\\
    \midrule
    Vehicle & Mechanical & $40$ & $360^{\circ}$ & $\le 200m$\\
    Infrastructure & Solid-state & $300$ & $100^{\circ}$ & $\le 280m$\\
  \bottomrule
\end{tabular}
\end{table}

\section{Experiments on More Simulated Datasets}

To further validate the effectiveness of DUSA, we select another simulated vehicle-to-vehicle collaborative perception dataset OPV2V \cite{opv2v} as the source domain and conducted more experiments. 
We implement DiscoNet \cite{disconet} (student only) as the collaborative detector and use the DAIR-V2X-C \cite{dair} dataset as the target domain. 
Other settings remain the same as the main paper.

\begin{table}
  \caption{Experiments on DiscoNet (student only) from the OPV2V dataset to the DAIR-V2X-C dataset. \textcolor{gray}{Gray texts} stand for the AP gain compared to \textit{No adaptation}.}
  \label{tab:supp-disco}
  \begin{tabular}{lccc}
    \toprule
    Method & AP @ 0.3 & AP @ 0.5 & AP @ 0.7 \\
    \midrule
    Oracle & 62.87 & 58.53 & 46.69 \\
    No adaptation & 37.28 & 33.79 & 20.24 \\
    \midrule
    Self-training & 32.79 \textcolor{gray}{(-4.49)} & 31.24 \textcolor{gray}{(-2.55)} & 22.02 \textcolor{gray}{(+1.78)} \\
    Discriminator & 42.04 \textcolor{gray}{(+4.76)} & 37.63 \textcolor{gray}{(+3.84)} & 22.17 \textcolor{gray}{(+1.93)} \\
    \textbf{DUSA (Ours)} & \textbf{43.83 \textcolor{gray}{(+6.55)}} & \textbf{39.59 \textcolor{gray}{(+5.80)}} & \textbf{24.06 \textcolor{gray}{(+3.82)}} \\
  \bottomrule
\end{tabular}
\end{table}

As \Cref{tab:supp-disco} shows, the performance of DUSA from the OPV2V dataset to the DAIR-V2X-C dataset is fairly good, demonstrating the effectiveness of DUSA among various simulated collaborative perception datasets.

\section{Extra Ablations on GRL Factors}

\begin{table}
  \caption{Ablation studies of the gradient reversal factors on V2X-ViT. $\gamma^{LSA}$ and $\gamma^{CIA}$ stand for the gradient reversal factors for the inputs of LSA and CIA respectively.}
  \label{tab:supp-ablations}
  \begin{tabular}{cc|ccc}
    \toprule
    $\gamma^{LSA}$ & $\gamma^{CIA}$ & AP @ 0.3 & AP @ 0.5 & AP @ 0.7\\
    \midrule
    $-0.025$ & $-0.1$ & 41.76 & 36.98 & 18.09\\
    $-0.05$ & $-0.1$ & \textbf{43.61} & \textbf{38.46} & \textbf{20.09}\\
    $-0.1$ & $-0.1$ & 40.68 & 35.79 & 17.05\\
    \midrule
    $-0.025$ & $-0.05$ & 38.37 & 33.70 & 15.43\\
    $-0.05$ & $-0.1$ & \textbf{43.61} & \textbf{38.46} & \textbf{20.09}\\
    $-0.1$ & $-0.2$ & 41.40 & 36.54 & 17.85\\
  \bottomrule
\end{tabular}
\end{table}

We conduct extra ablation studies on the gradient reversal factors of GRLs \cite{grl} before the LSA module and the CIA module. 
These factors rescale the reverse gradient flow to the feature extractor and balance the adversarial training process. 
The experiments are also done on the validation split of the DAIR-V2X-C \cite{dair} dataset with the V2X-ViT \cite{v2xvit} detector. 
As \Cref{tab:supp-ablations} shows, the gradient reversal factors have a prominent impact on the domain adaptation process and need to be appropriately selected.

\section{More Qualitative Results}

\begin{figure*}[ht]
  \centering
  \includegraphics[width=\textwidth]{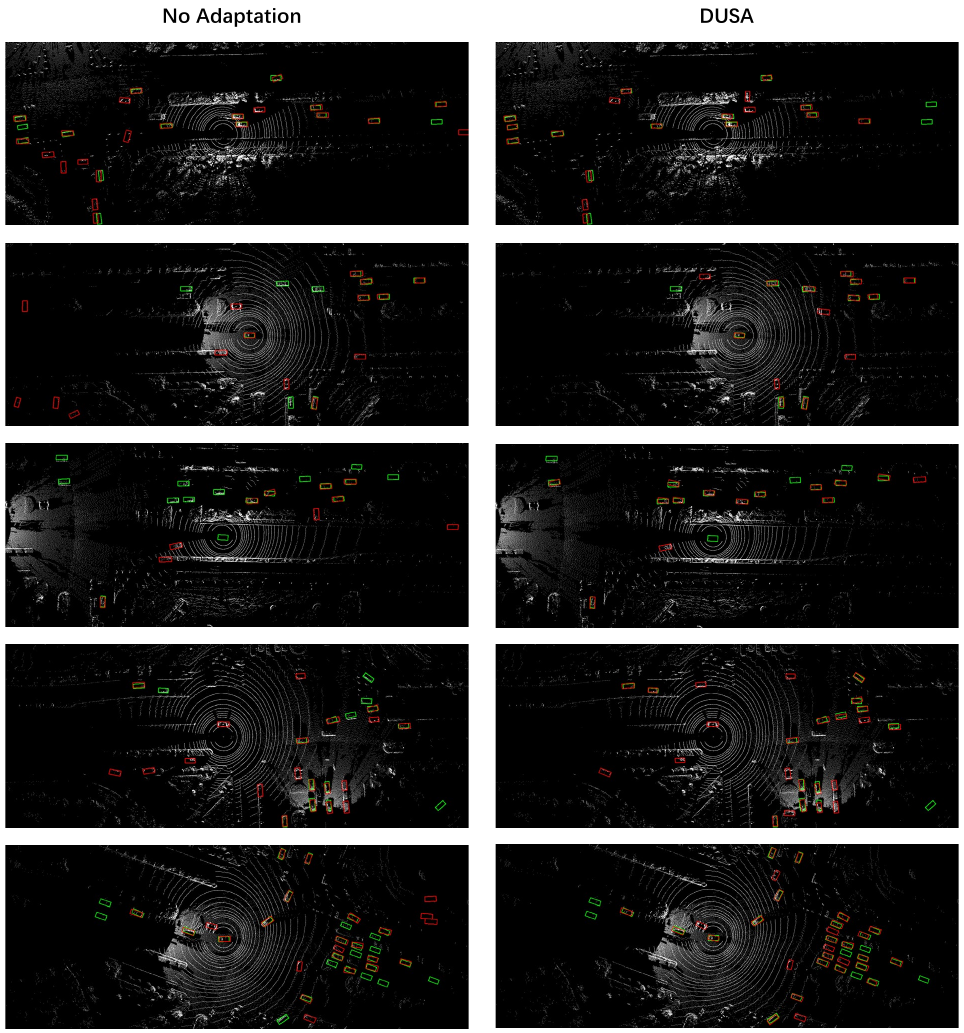}
  \caption{Visualization of detection results of the V2X-ViT model without domain adaptation and the DUSA-adapted V2X-ViT model. \textcolor{green}{Green boxes} are ground truths, and \textcolor{red}{red boxes} are predictions.}
  \Description{Qualitative results of DUSA.}
  \label{fig:vis-supp}
\end{figure*}

We provide more visualizations of the V2X-ViT \cite{v2xvit} model without domain adaptation and the DUSA-adapted V2X-ViT \cite{v2xvit} model on the validation split of the DAIR-V2X-C \cite{dair} dataset in \Cref{fig:vis-supp}.

\end{document}